\documentclass[journal]{IEEEtran}
\usepackage[letterpaper, left=54pt, right=54pt, bottom=54pt, top = 54pt]{geometry}
\newgeometry{left=54pt, right=54pt, bottom=54pt, top = 72pt}
\IEEEoverridecommandlockouts

\usepackage{amsmath}

\usepackage{amsfonts}
\usepackage{hyperref}
\usepackage{textcomp}
\usepackage{subcaption}

\usepackage{tikz}
\usepackage{xfrac}
\usepackage{siunitx}
\usepackage{float}
\usepackage{pgfplots}
\usepackage{tikz-3dplot}
\usetikzlibrary{calc}
\pgfplotsset{width=7cm,compat=1.8}

\usepackage{color}
\usepackage{soul}

\usepackage[style=ieee, citestyle=numeric-comp,sorting=none]{biblatex}
\def\BibTeX{{\rm B\kern-.05em{\sc i\kern-.025em b}\kern-.08em
    T\kern-.1667em\lower.7ex\hbox{E}\kern-.125emX}}

\newtheorem{remark}{Remark}
 
\begin{document}

\newcommand{\braces}[1]{\left\{#1\right\}}
\newcommand{\norm}[1]{\left\|#1\right\|_2}
\newcommand{\bs}[1]{\ensuremath{\boldsymbol{#1}}}

\newcommand{\Rp}{{\mathbf{R}^{gc}}}
\newcommand{\Rb}{{\mathbf{R}^{gb}}}
\newcommand{\Rr}{{\mathbf{R}^{cb}}}

\newcommand{\xg}{\ensuremath{\bs{x}^g}}
\newcommand{\eb}[1]{\ensuremath{\bs{e}^b_{#1}}}
\newcommand{\eg}[1]{\ensuremath{\bs{e}^g_{#1}}}
\newcommand{\esc}{\bs{e}^c_s}
\newcommand{\eyc}{\bs{e}^c_y}
\newcommand{\enc}{\bs{e}^c_n}

\newcommand{\xc}{\bs{x}^c}
\newcommand{\xcs}{\xc_s}
\newcommand{\xcsmag}{\norm{\xcs}}
\newcommand{\xcss}{\xc_{ss}}
\newcommand{\rc}{\bs{r}^c}
\newcommand{\rcs}{\rc_s}

\newcommand{\omb}{\ensuremath{\bs{\omega}^b}}
\newcommand{\omp}{\ensuremath{\bs{\eta}^p}}
\newcommand{\omr}{\ensuremath{\bs{\xi}^r}}
\newcommand{\Omb}{\ensuremath{\hat{\bs{\omega}}^b}}
\newcommand{\Omp}{\ensuremath{\hat{\bs{\eta}}^p}}
\newcommand{\Omr}{\ensuremath{\hat{\bs{\xi}}^r}}

\newcommand{\ks}{\kappa^c_s}
\newcommand{\ky}{\kappa^c_y}
\newcommand{\kn}{\kappa^c_n}

\newcommand{\vb}[1]{\ensuremath{v^b_{#1}}}
\newcommand{\wb}[1]{\ensuremath{\omega^b_{#1}}}

\title{Euclidean and non-Euclidean Trajectory Optimization Approaches for Quadrotor Racing\\
\thanks{Thomas Fork and Francesco Borrelli are with the Department of Mechanical Engineering, University of California, Berkeley, USA.}
\thanks{Corresponding author: Thomas Fork (fork@berkeley.edu).}
\thanks{Source code to be released to accompany final manuscript}
}

\author{\IEEEauthorblockN{Thomas Fork}
and
\IEEEauthorblockN{Francesco Borrelli}
}

\maketitle
\begin{abstract}
We present two quadrotor raceline optimization approaches which differ in using Euclidean or non-Euclidean geometry to describe vehicle position. Both approaches use high-fidelity quadrotor dynamics and avoid the need to approximate gates using waypoints. We demonstrate both approaches on simulated racetracks with realistic vehicle parameters where we demonstrate 100x faster compute time than comparable published methods and improved solver convergence. We then extend the non-Euclidean approach to compute racelines in the presence of numerous static obstacles. 

\end{abstract}


\section{Introduction}

Trajectory optimization for aircraft has received much interest for multirotors \cite{drone_min_snap} and fixed-wing aircraft \cite{red_bull_racing} with application in racing \cite{scaramuzza_racing_survey, airplane_racing}, acrobatics \cite{scaramuzza_deep_acrobatics}, interception \cite{mueller_drone_traj_opt} and navigation \cite{drone_navigation}. Racing poses the greatest demands on the performance, efficiency and accuracy of trajectory optimization, particularly for computing racelines: minimum-time, periodic trajectories through a racetrack.
Computing racelines is challenging on numerous fronts: nonconvexity due to nonlinear dynamics, progress that may be ill-defined between gates or waypoints, and potentially numerous static obstacles, all of which must be fully exploited for performance.

Numerous methods have been proposed to compute racelines. Early work focused on fixed-wing aircraft \cite{red_bull_racing,airplane_racing} while more recent work \cite{giuseppe_drone_racing, scaramuzza_cpc} has focused on multirotors. Two structural approaches have emerged: In \cite{red_bull_racing}, multiphase optimization is set up with a fixed allocation of finite elements to pairs of gates to pass through. In contrast, \cite{scaramuzza_cpc} determines allocation of finite elements online. \cite{scaramuzza_cpc} demonstrates performance exceeding a human expert in hardware experiments, however the half hour of compute time and need to treat gates as waypoints leave room for improvement.

In this paper we present two approaches for quadrotor trajectory optimization which improve upon \cite{scaramuzza_cpc} in key areas: We compute identical results 100x faster, avoid the approximation of gates as waypoints, and enable complex static obstacle avoidance. Our approaches differ in using Euclidean or non-Euclidean geometry, with the latter used for extension to obstacle avoidance. 

In the Euclidean approach, we adopt standard ideas from \cite{red_bull_racing} with several extensions, particularly to treat gates as objects to pass through instead of waypoints. Similar gate concepts have appeared in \cite{bos2022multi} which we apply to longer and periodic scenarios. Preprint publications \cite{zhou2023efficient} and \cite{qin2023time} have also improved upon the compute time of \cite{scaramuzza_cpc}. \cite{zhou2023efficient} requires treating gates as waypoints and solves an aperiodic problem. \cite{qin2023time} allows for flexible gate shape but is based on differential-flatness assumptions that result in conservative vehicle thrust constraints: the waypoint lap time of \cite{scaramuzza_cpc} is matched only when leveraging the full size of each gate.

In the non-Euclidean approach we change the coordinate system used to describe vehicle position. This is motivated by obstacle-rich environments where trajectory optimization requires an initial guess from search-based methods \cite{drone_search_planning, safe_flight_corridors}. This initial guess may be encoded by choice of geometry; such ideas have appeared for path-following \cite{arrizabalaga2022tunnels,ramirez2021gravity,scaramuzza_mpcc} and planning \cite{giuseppe_drone_racing}. We improve upon previous literature here in part by introducing geometry that is more general, captures prior literature as special cases, and avoids several previous mathematical errors. Following this, we develop an improved non-Euclidean trajectory optimization approach with subtle differences from the Euclidean approach due to the modified geometry. We show that this approach is similar in complexity to the first, with the key difference that the chosen geometry may be a limiting factor. 

We then augment our non-Euclidean approach with obstacle avoidance, the key advantage of this approach. This is an augmentation in the sense that the structure of the trajectory optimization problem is unchanged, and only requires adding convex constraints. This is in stark contrast to Euclidean approaches which involve a decomposition of free space into convex regions, known as a Safe Flight Corridor \cite{safe_flight_corridors}. Resulting trajectory optimization problems require additional optimization phases for a fixed order of regions \cite{planning_convex_decomp,safe_flight_corridors} or mixed-integer variables to encode the choice of region \cite{miqp_polyhedra_following,marcucci2024shortest}. These methods have shown impressive scaling \cite{marcucci2023fast} but only without dynamics or actuation limits. Our obstacle avoidance approach enables high-fidelity trajectory refinement even in the presence of many obstacles. 

This paper is structured as follows: In Section \ref{sec:preliminaries} we discuss related work on non-Euclidean geometry and introduce notation. We introduce vehicle kinematics in Section \ref{sec:kinematics} and dynamics in Section \ref{sec:dynamics}. We develop our trajectory optimization approaches in Sections \ref{sec:num_method} and \ref{sec:obstacle_avoidance} and apply them to simulated scenarios in Section \ref{sec:Results}.

\begin{figure}
    \centering
    \includegraphics[width=0.9\linewidth]{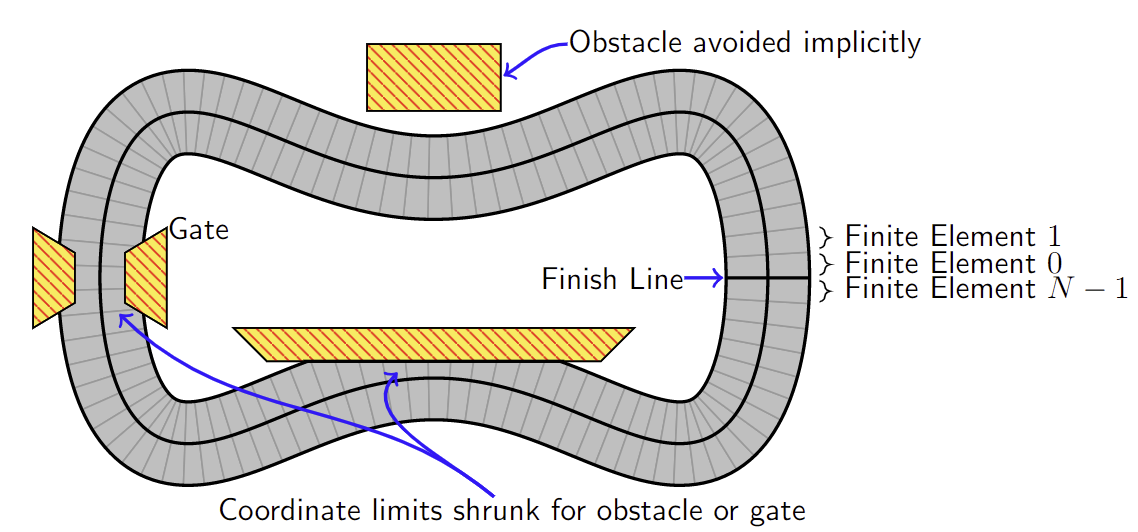}
    \caption{Schematic of a non-Euclidean coordinate system for racing. For trajectory optimization the racetrack is discretized along a curvilinear coordinate, with variable time for each finite element. Lateral coordinate limits of each element shrink to avoid obstacles and pass through gates.}
    \label{fig:race_schematic}
\end{figure}
\section{Preliminaries} \label{sec:preliminaries}

\subsection{Non-Euclidean Geometry Applied to Aircraft}
Non-Euclidean geometry such as illustrated in Figure \ref{fig:race_schematic} is widely used for racecars \cite{limebeer_car_racing_survey} but has only recently been used for aircraft.
In \cite{giuseppe_drone_racing} a curvilinear coordinate approach was introduced for quadrotor trajectory optimization. However,~\cite{giuseppe_drone_racing} contains geometric modeling errors which we discuss in Appendix \ref{app:giuseppe}. 
The work in \cite{arrizabalaga2022tunnels} avoids said issues but is limited to arc-length parameterizations of the curvilinear abscissa.
\cite{arrizabalaga2022spatially} extends \cite{arrizabalaga2022tunnels} to special cases of non-arc-length centerlines. However, their approach is ill-posed as the correlation between Euclidean and non-Euclidean position is lost (Appendix \ref{app:arrizabalaga}). For example, in Figure \ref{fig:race_schematic}, position at a given finite element is restricted to a line. \cite{arrizabalaga2022spatially} enforces this relationship only through the derivative of position, meaning integration errors accumulate and render pose kinematics incorrect. \cite{ramirez2021gravity} correctly introduces a gravity-referenced moving frame and allows for non-arc-length parameterized curves, but is not defined for vertical curves. 

In this paper we introduce a general non-Euclidean approach which overcomes the aforementioned issues and captures \cite{giuseppe_drone_racing,arrizabalaga2022tunnels,ramirez2021gravity} as special cases. 

\subsection{Notation} \label{sec:notation}
In the following we use subscripts to enumerate components and partial derivatives, with superscripts to denote frame of reference. For instance $\vb{1}$ is the $1$ component of vehicle velocity $\bs{v}$ in the $1$ direction of the body frame $b$, with velocity itself always w.r.t. an inertial frame.
We use lowercase for scalars and vectors with the latter in bold. Lowercase $\bs{e}$ is reserved for unit vectors. Uppercase bold symbols indicate matrices. We will at times treat vectors as a sum of basis components, that is $\bs{v}^b = \vb{1}\eb{1}+\vb{2}\eb{2}+\vb{3}\eb{3}$ as opposed to $\bs{v}^b= \left[\vb{1}~ \vb{2}~ \vb{3}\right]^T$. This will be clear based on usage, ie. the latter when multiplying by a matrix. Matrices will always be treated as algebraic.

We introduce a body-fixed frame of reference $\eb{1,2,3}$, and a global, inertial frame $\eg{1,2,3}$, both orthonormal. $\Rb$ denotes the rotation matrix such that $\left(\Rb\right)_{ij} = \eg{i} \cdot \eb{j} $.

We denote differentiation with respect to time by $\partial_t$, and exclusively consider time derivatives with respect to the inertial frame. We use $\dot{}$ to denote component-wise time rates, ie. $\dot{\bs{v}}^b=\sum_{i=1,2,3}\dot{v}^b_i \eb{i}$ while $\partial_t \bs{v}^b = \dot{\bs{v}}^b + \sum_{i=1,2,3}\vb{i} \partial_t \eb{i}$.

We use $\hat{~}$ for the the skew-symmetric operator, ie. $\Omb$ is:
\begin{equation} \nonumber
    \Omb = 
    \begin{bmatrix}
        0 & -\wb{3} & \wb{2} \\ \wb{3} & 0 & -\wb{1} \\ -\wb{2} & \wb{1} & 0
    \end{bmatrix}.
\end{equation}

\section{Vehicle Kinematics} \label{sec:kinematics}

\subsection{Euclidean Kinematics} \label{sec:global_kinematics}
Euclidean pose kinematics are fully described by:
\begin{align}
    \partial_t \xg &= \Rb \bs{v}^b &
    \partial_t \Rb &= \Rb \Omb. \label{eq:euclidean_orientation_kinematics}
\end{align}
Here $x$ and $\omega$ are the position and angular velocity of the body frame with respect to the global frame. It is common practice to replace the nine components of $\Rb$ with Euler angles or a quaternion. In all cases, \eqref{eq:euclidean_orientation_kinematics} can be replaced with one of the form $\dot{\bs{r}} = \bs{M}(\bs{r}) \omb$, where $\bs{r}$ is a vector of Euler angles or quaternion components and $\bs{M}(\bs{r})$ a matrix function of $\bs{r}$. These equations are standard and omitted here.

\subsection{Non-Euclidean Kinematics} \label{sec:parametric_kinematics}
\begin{figure}[ht]
    \centering
    \includegraphics[width=0.8\linewidth]{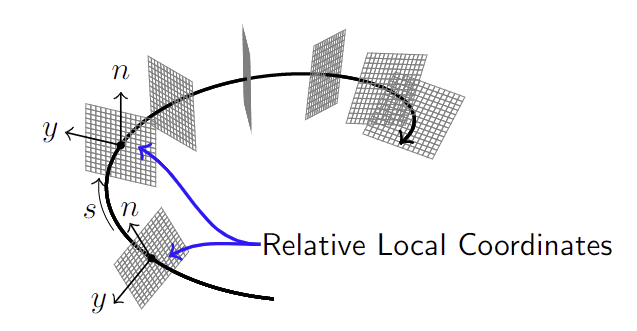}
    \caption{Non-Euclidean coordinate system. A curve parameterized by $s$ is augmented by a frame that moves along this curve, with relative coordinates $y$ and $n$.}
    \label{fig:relative_coords}
\end{figure}

We introduce an approach for expressing pose kinematics in a frame of reference that moves along a curve, ie. the frame itself is a function of position along the curve. This is illustrated in Figure \ref{fig:relative_coords} where variable $s$ parameterizes the curve and variables $y$ and $n$ are relative position; $y-n$ planes at fixed $s$ are Euclidean and orthogonal to the tangent of the curve.

We refer to the curve as the centerline and introduce the ``parametric frame" defined by the centerline $\xc$ and orthonormal basis vectors $\esc$, $\eyc$ and $\enc$ as shown in Figure \ref{fig:coordinate_systems}. This provides a natural way to express pose along and relative to a curve, such as an initial trajectory guess.
We do not assume $s$ is an arc length parameterization of the centerline. Next we derive motion equations and restrictions in the parametric frame.

\subsubsection{Position Kinematics} \label{sec:pose_kinematics}

\begin{figure}
    \centering
    \includegraphics[width=0.7\linewidth]{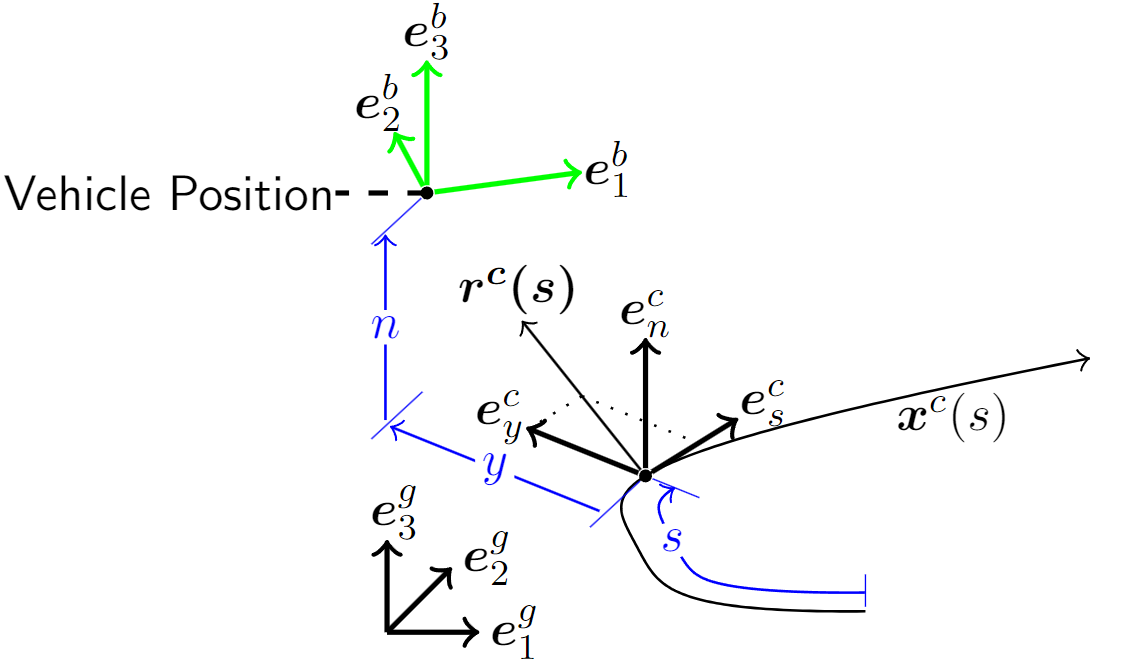}
    \caption{Frames of reference used in our approach. Superscript $g$ denotes an inertial Euclidean frame of reference also referred to as the global frame. Superscript $b$ denotes the body-fixed Euclidean frame and $c$ denotes a non-Euclidean frame which varies along the centerline $\xc$. Note that $\xc(s)$ may not be an arc length parameterization of the centerline.}
    \label{fig:coordinate_systems}
\end{figure}

In this section we derive kinematics relating vehicle velocity to $\dot{s}$, $\dot{y}$ and $\dot{n}$. We do so by expressing position $\bs{x}$ in terms of $s$, $y$ and $n$ and differentiating with respect to time. We start with position:
\begin{equation} \label{eq:xg}
    \bs{x} = \xc(s) + y \eyc(s) + n \enc(s),
\end{equation}
where  as illustrated in Figure \ref{fig:coordinate_systems}, $\xc(s)$ is a curve in $\mathbb{R}^3$ with tangent, binormal and normal vectors $\esc$, $\eyc$ and $\enc$ respectively. When $\xc$ is an arc length parameterization we have $\xcsmag = 1$, where $\xcs = \partial_s\xc$. This is a case of the Darboux frame \cite{differential_geometry_of_curves_and_surfaces}, where by definition:
\begin{subequations} \label{eq:darboux_frame}
    \begin{align}
        \partial_s \esc &= \kn \eyc + \ky \enc \\
        \partial_s \eyc &= \ks \enc - \kn \esc \\
        \partial_s \enc &= -\ky \esc - \ks \eyc.
    \end{align}
\end{subequations}
Here $\ks$, $\ky$ and $\kn$ are curvature terms about their respective basis vectors, known as torsion\footnote{$\ks = 0$ is the torsion-free case and is known as the Frenet-Serret frame.}, normal curvature, and geodesic curvature respectively. This frame has convenient expressions for $\dot{s}$, $\dot{y}$ and $\dot{n}$, but requires sophisticated curve-fitting to apply to a known 3D shape \cite{3d_part_1}. We bridge this gap by assuming that $\xc(s)$ may not be an arc length parameterization and that we only know or have chosen $\rc(s)$, a vector close to $\eyc$ but which may not be orthonormal to $\xcs$, as shown in Figure \ref{fig:coordinate_systems}. This allows $\xc$ to be fit to known spatial waypoints with simple interpolation methods such as a spline. Flexible choice of $\rc$ allows for the creation of coordinate systems that hold special meaning, such as being torsion-free or having a relationship to gravity \cite{ramirez2021gravity}.  We reduce our general case to \eqref{eq:darboux_frame}, beginning with orthonormalization of $\xcs$ and $\rc$. Denoting projection to unit 2 norm with $\sim$ instead of $=$ we have:
\begin{subequations} \label{eq:induced_darboux_frame}
    \begin{align}
        \esc &\sim \xcs \\
        \eyc &\sim \rc - \esc\left(\esc\cdot\rc\right)\\
        \enc &= \esc \times \eyc.
    \end{align}
\end{subequations}

By change of variables, when $\xcsmag\neq 1$, \eqref{eq:darboux_frame} becomes:
\begin{subequations} \label{eq:darboux_time_diff}
    \begin{align}
        \partial_t \esc &= \left(\kn \eyc + \ky \enc\right) \xcsmag \dot{s} \\
        \partial_t \eyc &= \left(\ks \enc - \kn \esc\right) \xcsmag \dot{s} \\
        \partial_t \enc &= \left(-\ky \esc - \ks \eyc\right) \xcsmag \dot{s}.
    \end{align}
\end{subequations}
We use this to derive expressions for $\ks$, $\ky$ and $\kn$ from $\xc$, $\rc$ and their partial derivatives with respect to $s$.

First, as the parametric frame is orthogonal, we have $\enc\cdot\xcs=\eyc\cdot\xcs=0$ at all times. Therefore, $\partial_t\left(\enc \cdot \xcs\right) = \partial_t\left(\eyc \cdot \xcs\right) = 0$. Expanding terms, we have
\begin{subequations}
    \begin{align}
        \left(-\ky \esc - \ks \eyc\right) \cdot \xcs \xcsmag \dot{s} + \enc \cdot \xcss \dot{s} &= 0 \\
        \left(-\ky \esc - \ks \eyc\right) \cdot \rc  \xcsmag \dot{s} + \enc \cdot \rcs  \dot{s} &= 0,
    \end{align}
\end{subequations}
which results in:
\begin{equation} \label{eq:ksky}
    \begin{bmatrix}
        \ky \\ \ks
    \end{bmatrix}
    =
    \begin{bmatrix}
        \xcs \cdot \esc & \xcs \cdot \eyc \\
        \rc  \cdot \esc  & \rc \cdot \eyc 
    \end{bmatrix}^{-1}
    \begin{bmatrix}
        \enc \cdot \xcss \\ \enc \cdot \rcs
    \end{bmatrix}
    \frac{1}{\xcsmag}.
\end{equation}
The inverse exists except for pathological choices of $\xc$ and $\rc$, such as $\rc$ parallel to $\xcs$ or either being 0. All terms result directly from smooth interpolation of $\xc$ and $\rc$.

For $\kn$ we manipulate \eqref{eq:darboux_time_diff} such that: 
\begin{subequations}
    \begin{align}
        \partial_t \esc =& \left(\kn\eyc + \ky\enc\right)\xcsmag \dot{s}\\
        \partial_t \esc \times \esc =& \left(-\kn \enc + \ky \eyc\right) \xcsmag \dot{s} \\
        \left(\partial_t \esc \times \esc\right) \cdot \enc =& -\kn \xcsmag \dot{s}. \label{eq:kn_expr_1}
    \end{align}
\end{subequations}
$\esc$ is known from \eqref{eq:induced_darboux_frame}, from which $\partial_t \esc$ follows:
\begin{equation}
    \partial_t \esc = \frac{\xcss \dot{s}}{\xcsmag} + \xcs( \dots).
\end{equation}
The omitted $\xcs$ term vanishes in \eqref{eq:kn_expr_1} and
\begin{equation} \label{eq:kn}
    \kn = -\frac{\left(\xcss \times \xcs\right)\cdot \enc}{\xcsmag^3}.
\end{equation}

Equations \eqref{eq:ksky} and \eqref{eq:kn} fill in the necessary terms for \eqref{eq:darboux_time_diff}. 
We complete this section by differentiating \eqref{eq:xg} with respect to time to determine equations for $\dot{s}$, $\dot{y}$ and $\dot{n}$. 
The time derivative of $\bs{x}$ is velocity, the remainder follows from \eqref{eq:darboux_time_diff} and standard calculus.
\begin{equation}
\begin{split}
    \bs{v} &= \xcs \dot{s} + \dot{y} \eyc + \dot{n} \enc \\
           &+ y \left(\ks \enc - \kn \esc \right) \xcsmag \dot{s} \\
           &+ n \left(-\ky \esc - \ks \eyc \right) \xcsmag \dot{s}.
\end{split}
\end{equation}
We factor this into three equations by taking inner products with respect to $\esc$, $\eyc$ and $\enc$:
\begin{subequations} \label{eq:position_kinematics}
    \begin{align}
        \dot{s} &= \frac{\bs{v} \cdot \esc}{\left(1 - \ky n - \kn y \right)\xcsmag} \label{eq:s_kinematics}\\
        \dot{y} &= \bs{v} \cdot \eyc + n \ks \xcsmag \dot{s} \\
        \dot{n} &= \bs{v} \cdot \enc - y \ks \xcsmag \dot{s}.
    \end{align}
\end{subequations}
Equation \eqref{eq:position_kinematics} completes the position equations of motion. The components of $\bs{v}$ needed for \eqref{eq:position_kinematics} follow from vehicle orientation. It is necessary that the denominator of \eqref{eq:s_kinematics} be nonzero, this is discussed in Section \ref{sec:kinematic_remarks}.

\subsubsection{Orientation Kinematics} \label{sec:orientation_kinematics}
It is common practice to express vehicle orientation by a rotation between the global frame and body frame, such as $\Rb$. The presence of the parametric frame allows for a second option, a rotation between the parametric frame basis vectors $\esc,~\eyc$ and $\enc$ and the body frame: $\Rr$. The two choices are interchangeable, albeit with different kinematics. We introduce these here.

First, it is immediate that:
\begin{subequations}
    \begin{align}
    \Rb &= \Rp \Rr\\
    \partial_t \Rb &= \Rb \Omb\\
    \partial_t \Rp &= \Rp \Omp,
    \end{align}
\end{subequations}
with $\omp = \left[\ks\ -\ky\ \kn \right]^T \xcsmag \dot{s}$ per \eqref{eq:darboux_time_diff}. It follows from standard rotation theory \cite{murray_robotics} that the effective angular velocity $\omr$ on $\Rr$ is:
\begin{equation}
    \omr = \omb - (\Rr)^T\omp,
\end{equation}
which factors into any parameterization of $\Rr$.

Given their relationship, usage of $\Rr$ instead of $\Rb$ is completely optional. $\Rr$ is highly sensitive to the chosen parametric frame of reference and in practice we have not found it useful for trajectory optimization. 

\subsection{Remarks on Non-Euclidean Geometric Approach} \label{sec:kinematic_remarks}
We conclude this section with several important remarks:

\begin{remark} \label{remark:fixed_s}
    Frame-dependent terms enter the dynamics of a dynamical system with non-Euclidean pose. In our approach these terms are $\xc$, $\xcs$, $\xcss$, $\rc$ and $\rcs$, and are functions of $s$. These terms become constant when we fix finite elements in $s$ (Figure \ref{fig:race_schematic}), simplifying trajectory optimization.
\end{remark}

\begin{remark}
    By linking $\xc$ and $\rc$ to the Darboux frame our approach can be applied directly to an environment, eg. by fitting a cubic spline to waypoints. This improves upon approaches that require an arc-length parameterization for the centerline \cite{arrizabalaga2022tunnels} and Darboux frame coefficients $\ks$, $\ky$ and $\kn$. These are difficult to fit to known geometry, for example \cite{3d_part_1} used optimization to fit a Darboux frame to 3D roads. We also capture previous literature as special cases, for example \cite{ramirez2021gravity} is the case where $\rc$ is chosen orthogonal to $\xcs$ and the direction of gravity. The best choice of geometry is bound to be application-dependent.
\end{remark}

\begin{remark}
    The choice of centerline is an inherent limiting factor. Mathematically, \eqref{eq:s_kinematics} becomes singular at certain combinations of curvature and lateral position. Geometrically, were we to extrude a circle along $\xc$ thus forming a tube, these singularities correspond to where the sides of the tube intersect such that moving along $s$ causes no change in position. To avoid this during trajectory optimization, we enforce the constraint:
    \begin{equation} \label{eq:regularity_limit}
         \ky n + \kn y \leq \lambda < 1,
    \end{equation}
    where $\lambda$ determines how close to the singularity it is acceptable to move. This is a consequence of the coordinate system alone - clearly vehicle motion is unchanged but differential equations for $s$, $y$ and $n$ do not exist if \eqref{eq:regularity_limit} is violated. \cite{arrizabalaga2022spatially} introduced an approach to avoid this but is incorrect (Appendix \ref{app:arrizabalaga}).
\end{remark}

\begin{remark}
    Equation \eqref{eq:regularity_limit} prevents any claims of global optimality. We make no claim as to the ``best" choice of centerline and consider this an area of possible future research.
\end{remark}

\begin{remark}
    Ensuring dynamic feasibility of a trajectory is unchanged from Euclidean geometry as vehicle motion is preserved but described in a new coordinate system.
\end{remark}

\section{Vehicle Dynamics} \label{sec:dynamics}

Vehicle dynamics require modeling how linear and angular velocity change as a function of pose, velocity and other system inputs. As we express velocity in the body frame, the corresponding rigid body equations are~\cite{mechanics_landau_lifshitz}:
\begin{align}\label{eq:equations_of_motion}
        m\left(\dot{\bs{v}}^b + \Omb \bs{v}^b \right) &= \bs{F}^b
        &
        \bs{I}^b \dot{\bs{\omega}}^b + \Omb \bs{I}^b \omb  &= \bs{K}^b.
\end{align}
$\bs{F}^b$ and $\bs{K}^b$ are net body frame force and moment respectively. For instance the body frame forces due to gravity are the components thereof of $-mg\eg{3}$. We discuss vehicle-specific examples next.

\subsection{Point Mass}
A point mass by definition has no inherent notion of orientation; we treat it as an oriented point mass instead, with the global frame used as the body frame\footnote{Second order equations of motion for $s$, $y$ and $n$ result from another round of differentation akin to Section \ref{sec:pose_kinematics}, with which the notion of a ``body frame" can be removed. Equation 2.13 of \cite{intro_relativity} is the general result, but the lack of a body frame precludes use for aircraft.}. We add an input force vector with magnitude bounded by the thrust to weight ratio (TWR) of the vehicle which acts in addition to gravity on the point mass.

\subsection{Quadrotor}
We use the four propeller thrust model of \cite{scaramuzza_cpc} with rotor thrusts $T_1$ through $T_4$:
\begin{subequations} \label{eq:quad_forces}
    \begin{align}
        F^b_1 &= -mg\eg{3} \cdot \eb{1} \\
        F^b_2 &= -mg\eg{3} \cdot \eb{2} \\
        F^b_3 &= T_1 + T_2 + T_3 + T_4 -mg\eg{3} \cdot \eb{3} \\
        K^b_1 &= \left(T_1 + T_2 - T_3 - T_4 \right) l \\
        K^b_2 &= \left(-T_1 + T_2 + T_3 - T_4 \right) l \\
        K^b_3 &= \left(T_1 - T_2 + T_3 - T_4 \right) c_\tau.
    \end{align}
\end{subequations}
Here $l$ is a geometric factor for the size of the quadrotor and $c_\tau$ is a propeller drag torque factor. We limit thrusts to the interval $\left[T_{min}, \frac{1}{4}\text{TWR}\right]$ with $T_{min}$ a model parameter. Net force and moment bounds are emergent from \eqref{eq:quad_forces}.

\subsection{Other Rigid-Body Vehicles}
Though we focus on the previous two models, other models may be incorporated within our trajectory optimization approaches by appropriate choice of dynamic equations and inputs, such as fixed wing-aircraft \cite{airplane_racing}. 

\begin{figure}
    \centering
    \includegraphics[width=\linewidth]{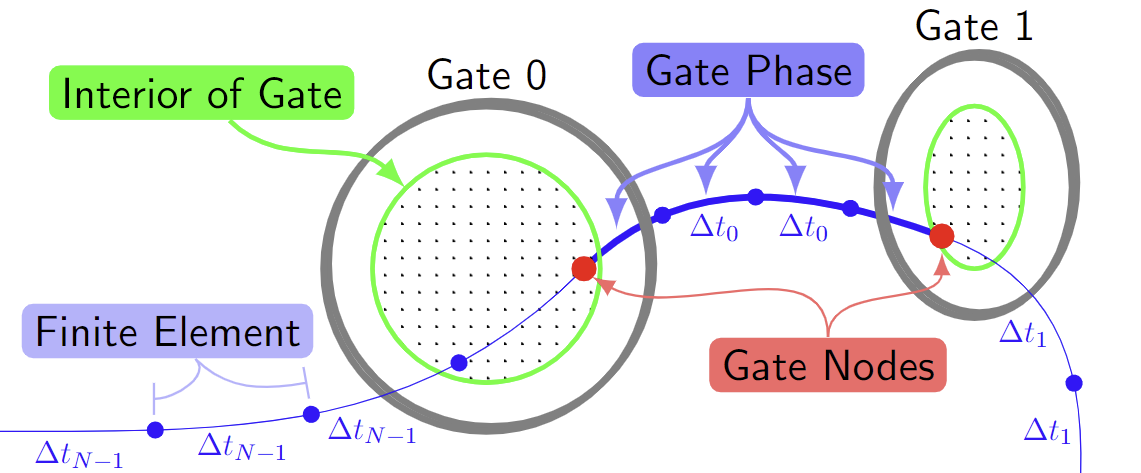}
    \caption{Euclidean Approach: Phases group finite elements from each gate to the next. Gate interior constraints at either end are shrunk for the diameter of the drone. Each element of a phase is given the same, variable time duration.}
    \label{fig:global_opt_method}
\end{figure}

\section{Trajectory Optimization Approaches} \label{sec:num_method}
We present the proposed trajectory optimization approaches. These differ in using Euclidean (Section \ref{sec:global_kinematics}) or non-Euclidean (Section \ref{sec:parametric_kinematics}) kinematics, and both address the task of computing a periodic, minimum-time raceline through a set of ordered gates. We discuss the structure of the multiphase optimization problem and state constraints imposed by each approach. We then augment the non-Euclidean approach with obstacle avoidance in Section \ref{sec:obstacle_avoidance}.
\subsection{Euclidean Trajectory Optimization Approach} \label{sec:global_planning}

In the global frame there is no notion of progress apart from gates or waypoints to pass through. This has been approached by either assigning finite elements to each waypoint \cite{red_bull_racing} or recomputing waypoint allocation during optimization \cite{scaramuzza_cpc}. The latter can be viewed as a mixed-integer problem though \cite{scaramuzza_cpc} does not formulate the problem as such. We propose an approach similar to \cite{red_bull_racing} with the addition of gate constraints and application to quadrotor dynamics.

We set up a multiphase problem \cite{gpops_ii} where each gate to pass through corresponds to a phase. To this end, several finite elements are assigned to an individual phase, with the endpoints of each phase fixed appropriately in space by adding state constraints to the optimization problem. For instance a circular gate is implemented by constraining vehicle position to a disk of smaller diameter, shrunk to capture the size of the vehicle as shown in Figure \ref{fig:global_opt_method}. Any convex shape is suitable and unlike \cite{scaramuzza_cpc} we can solve for trajectories that pass directly through a waypoint. The time duration of all finite elements within a phase are kept identical to avoid nonunique minima for these variables. Setup is determined by the order, location, and size of gates, the finite elements and numerical method, and the vehicle model.

In Euclidean approaches gate behaviour may be unpredictable: trajectories may enter a gate and backtrack without exiting the other side. While this is optimal for certain scenarios, it may not be desirable or allowed in practice. For the simulated racetracks studied in this paper, our method did not require additional considerations to prevent backtracking, but compatible methods for this have appeared in \cite{arrizabalaga2022spatially, bos2022multi, ramos2021minimum}.

\subsection{Non-Euclidean Trajectory Optimization Approach} \label{sec:parametric_planning}
\begin{figure}
    \centering
    \includegraphics[width=\linewidth]{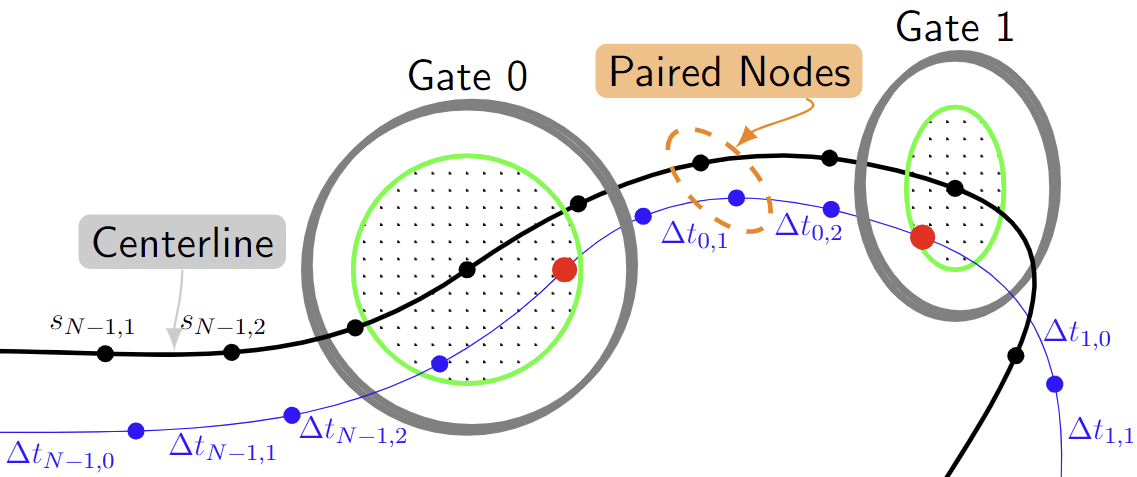}
    \caption{Non-Euclidean Approach: Finite elements are given fixed $s$ coordinate corresponding naturally to points along the centerline. Every finite element is given its own time duration and regularity constraints.}
    \label{fig:param_opt_method}
\end{figure}
With the use of non-Euclidean geometry there is a natural notion of progress along a racetrack, even without gates. Behaviour at gates is determined by the centerline, such as passing through a gate or entering and then backtracking.

We set up a similar multiphase problem to the Euclidean approach, with several key differences. Finite elements were given fixed $s$ coordinate along the centerline as illustrated in Figure \ref{fig:param_opt_method}. This simplifies geometric effects on vehicle dynamics (Remark \ref{remark:fixed_s}). Every finite element is fixed separately with independent, variable time duration. Gate constraints were enforced as constraints on $y$ and $n$, with consideration of the orientation of $\eyc$ and $\enc$ relative to the gate during setup. Depending on where finite elements are fixed, gates may lie within finite elements, not just at their endpoints. This may be approached by imposing constraints upon an interpolated state within the element. It may also be avoided entirely by choice of centerline parameterization and where elements are fixed, such as by choosing $s=k$ and gate number $k$ and assigning a positive number of elements to each gate transition, much like the Euclidean approach.

\begin{figure}
    \centering
    \includegraphics[width=\linewidth]{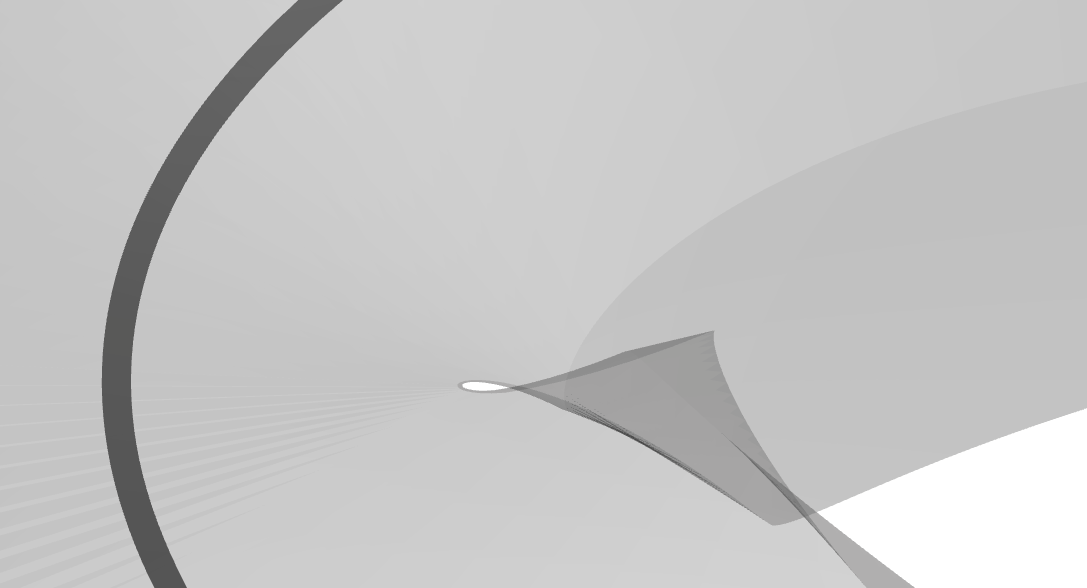}
    \caption{Regularity limits (Eqn. \ref{eq:regularity_limit}) illustrated by extruding a translucent gray tube along the centerline, colored black. Lateral limits of the tube are shrunk for regularity limits with $\lambda = 0.9$, causing the small opening at center. Any non-Euclidean trajectory must pass to the left of this opening.}
    \label{fig:regularity}
\end{figure}

In non-Euclidean approaches, limitations emerge from the choice of geometry. These appear in our approach through regularity constraints, Equation \ref{eq:regularity_limit}, required to avoid kinematic singularities. This is a source of conservativeness as trajectories are restricted to a neighborhood of the centerline when it is curved. Figure \ref{fig:regularity} illustrates the resulting limitations, which may impact performance depending on the size and location of gates. 

\subsection{Implementation Remarks} \label{sec:planning_remarks}
Both approaches fall within standard nonlinear multiphase optimization techniques such as \cite{gpops_ii} and general purpose solvers such as \cite{ipopt}. Tailored optimization routines may improve performance \cite{vanroye2023fatrop} but application to problems with periodic constraints remains an open problem.

In Section \ref{sec:Results} we implement all approaches with a fixed number of finite elements, which were determined manually. However, automatic refinement algorithms such as in \cite{gpops_ii} may automate this process.

\clearpage

Raceline optimization is nonconvex due to nonlinear drone dynamics and the variable time duration of finite elements. Sufficiently poor initial guesses will invariably lead to convergence of optimization algorithms to local, not global minima. To approach this in a systematic manner we set up a procedure to warmstart drone racelines by first solving for a point mass raceline with identical thrust to weight ratio. This was used to warmstart the pose and speed of the drone model, with drone orientation chosen to align the propellers with the point mass thrust. 

\section{Stationary Obstacle Avoidance with Non-Euclidean Approach} \label{sec:obstacle_avoidance}

Motion planning with both static obstacle avoidance and vehicle dynamic limits is a fundamentally challenging problem. We consider the common approach where an initial collision-free trajectory is obtained first, such as by a search-based method \cite{aggarwal_path_planning_survey, scaramuzza_waypoint_search}. Then, this trajectory is locally optimized to satisfy vehicle dynamics while retaining collision avoidance. We propose an approach for obstacle avoidance with the non-Euclidean approach to enable this.

We do so by recognizing that in the non-Euclidean approach, vehicle position is fixed to a 2D plane at finite elements. Local, convex, and collision free regions of this plane may be computed and imposed as collision-avoidance constraints during trajectory optimization. This adds no new phases or structural changes to the trajectory optimization problem, in contrast to requiring mixed-integer variables \cite{richards2002aircraft,miqp_polyhedra_following}, additional variables \cite{obca} or additional phases. Euclidean approaches require additional phases to plan through a sequence of collision-free convex regions \cite{planning_convex_decomp}, often referred to as a safe flight corridor (SFC) \cite{safe_flight_corridors}. Every transition between regions requires a new optimization phase, with a structure similar to our Euclidean approach. However, whereas the number of gates is fixed, many convex regions are required to accurately represent a nonconvex shape such as a u-turn.

\begin{figure}
    \centering
    \includegraphics[width=\linewidth]{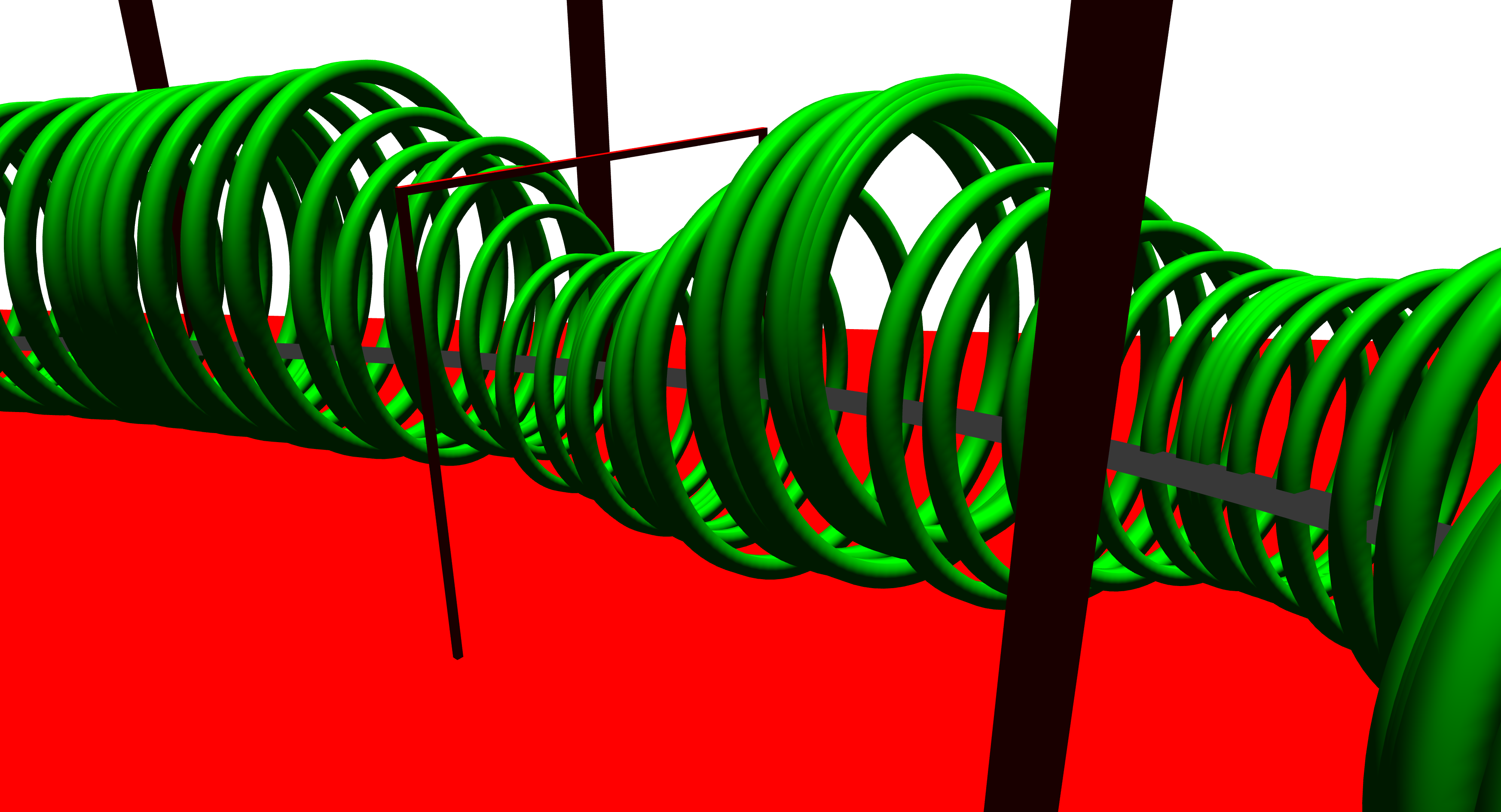}
    \caption{Illustration of safe-flight corridor (green rings) computed in an obstacle-rich environment. The centerline (black) does not need to be collision-free, as evidenced where the centerline is outside of the SFC. The full SFC is shown in Figure \ref{fig:r3_tube}.}
    \label{fig:sfc_example}
\end{figure}

Our method is in essence a SFC in non-Euclidean coordinates, where we leverage the coordinates themselves to encode obstacle avoidance and progress through an environment. This is illustrated in Figure \ref{fig:sfc_example}, where non-Euclidean coordinate limits are illustrated by a sequence of rings, with the interior of each ring corresponding to collision-free positions for a particular phase. Collision avoidance is immediate, as in safe flight corridors, but no new optimization phases have been introduced, only convex constraints on existing phases. This holds independent of the size or complexity of the environment, though the number of phases needed will be environment-dependent. In the case of a spherical object, often used to approximate quadrotors, $y$ and $n$ alone are constrained. Other vehicle shapes require considering vehicle orientation or approximating the vehicle as an enlarged sphere.

Additional considerations are necessary with the collision-free cross sections used. For instance, disjoint neighboring sections would be impossible for a vehicle to cross. Similarly, the choice of convex region to use may be unclear, such as when the original collision-free cross section is L-shaped. We consider these an area of potential future work and implement a simplified approach to illustrate our method. We run gridded search over $y$ and $n$ at every fixed $s$ coordinate along the centerline. Then, the point furthest from any obstacle is used for a circular collision-free region. Future work may explore faster algorithms and exploit the parallel nature of the problem. Our implementation takes roughly $13\unit{\second}$ to compute 800 circular regions for the example in Section \ref{sec:r3}, where we demonstrate our approach with trajectory optimization through a complex static obstacle field.

Similar methods are widespread with ground vehicles and have appeared for aircraft in \cite{giuseppe_drone_racing}. Unlike both, we can choose a SFC that does not contain the centerline, meaning we can improve upon the obstacle avoidance threshold used to obtain a centerline (Figure \ref{fig:sfc_example}) in addition to refining simplified vehicle models used to obtain an initial trajectory guess.

\clearpage
\section{Results} \label{sec:Results}

We applied our trajectory optimization approaches to several quadrotor racing scenarios and compared to \cite{scaramuzza_cpc}. Comparison to other methods was not pursued due to differences in vehicle model fidelity \cite{qin2023time,flatness_trajectory_opt} or the problem being solved, ie. aperiodic trajectories \cite{ramos2021minimum, giuseppe_drone_racing} and path-following \cite{arrizabalaga2022tunnels, ramirez2021gravity}.

We used the point mass and quadrotor drone models of Section \ref{sec:dynamics} with our Euclidean (Section \ref{sec:global_planning}) and non-Euclidean (Section \ref{sec:parametric_planning}) trajectory optimization approaches. We interchangeably refer to these as global and parametric. We used model parameters identical to \cite{scaramuzza_cpc} (Appendix \ref{app:parameters}). In all cases we expressed vehicle velocity in the body frame and orientation relative to the global frame. Where not stated otherwise, vehicle orientation was expressed by a quaternion and a point mass model raceline was used to warmstart drone raceline optimization as discussed in Section \ref{sec:planning_remarks}. When using quaternions, a projection to unit norm was added to continuity constraints between finite elements. Regularity constraints \eqref{eq:regularity_limit} were enforced with $\lambda=0.9$.

All optimization problems were set up using CasADi \cite{Andersson2018} version 3.5.5 in Python 3.9 using direct orthogonal collocation \footnote{An introduction to direct orthogonal collocation may be found in \cite{gpops_ii} and several related methods are discussed in \cite{kelly2017introduction}}. We used 7\textsuperscript{th} order Gauss-Legendre collocation with the number of intervals reported separately for each scenario. Optimization problems were solved by IPOPT \cite{ipopt} using the linear solver MUMPS \cite{MUMPS_1}. Code was run on an 11\textsuperscript{th} Gen Intel\textregistered\ Core\textsuperscript{TM} i7-11800H @2.3GHz.

We report lap time, solve time and setup time for all racelines, where setup time corresponds to the time required by CasADi to set up expressions for gradients and Hessians of the optimization problem. We further split solve time into the time spent in IPOPT and the time required for functions evaluations, labeled ``IPOPT" and ``feval" in subsequent tables. Timing statistics for point mass racelines used as a warmstart are reported separately.

\subsection{Racetrack Scenario} \label{sec:r1}

\begin{figure}
    \centering
    \includegraphics[width=\linewidth]{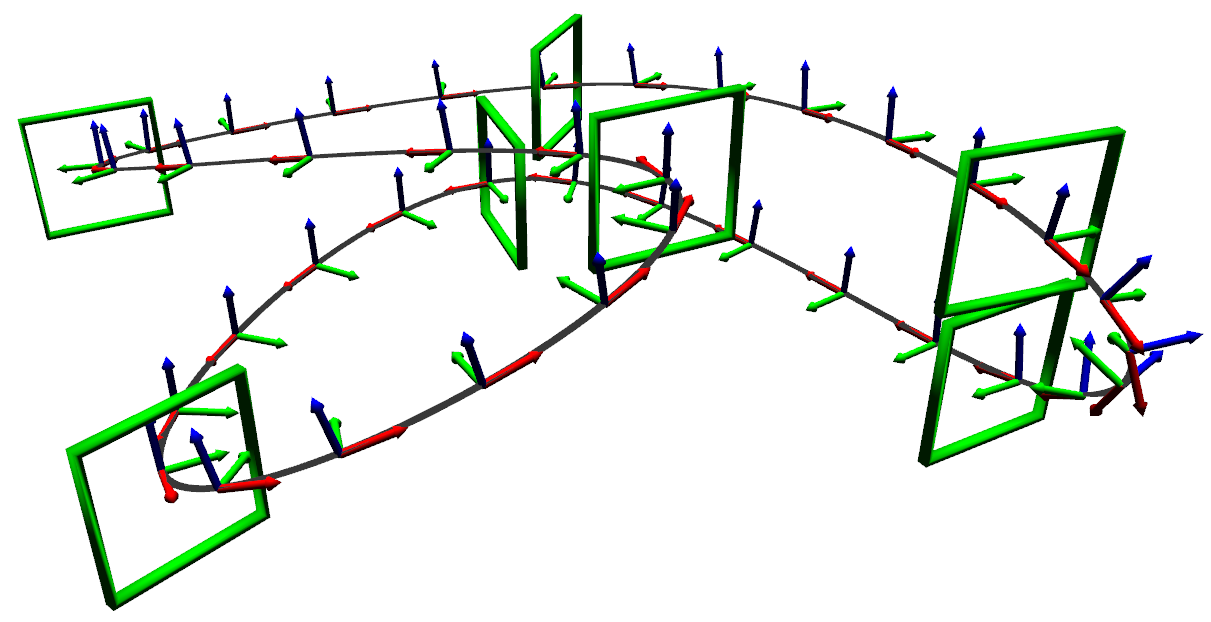}
    \caption{Illustration of the race scenario for Section \ref{sec:r1}, with gate locations adapted from \cite{scaramuzza_cpc}. Squares represent gates which a quadrotor must pass through in order. The parametric frame used is shown with centerline colored black and $\esc$, $\eyc$ and $\enc$ colored red, green and blue respectively. }
    \label{fig:r1_scenario}
\end{figure}

We first compared CPC \cite{scaramuzza_cpc} and our trajectory optimization approaches on the race scenario of \cite{scaramuzza_cpc}. We set up a parametric frame of reference for our non-Euclidean approach using a periodic cubic spline fit to the waypoints used in \cite{scaramuzza_cpc}. We chose $\rc$ waypoints at each gate that were orthogonal to the centerline tangent and lay within the horizontal plane. Between gates $\rc$ was interpolated with a periodic cubic spline and the parametric frame derived as described in Section \ref{sec:parametric_kinematics}. This is similar in nature to \cite{ramirez2021gravity}, but handles the vertical u-turn without issue which cannot be handled by \cite{ramirez2021gravity}. Both the centerline and parametric frame basis vectors are shown in Figure \ref{fig:r1_scenario}. Gates are shown with a square shape and inner width of 2.5$\unit{\meter}$. The drone was treated as a sphere with radius 0.3$\unit{\meter}$ for collision avoidance with the gates. Ten collocation intervals were allocated to each pair of gates.

\begin{figure}
    \centering
    \includegraphics[width = \linewidth]{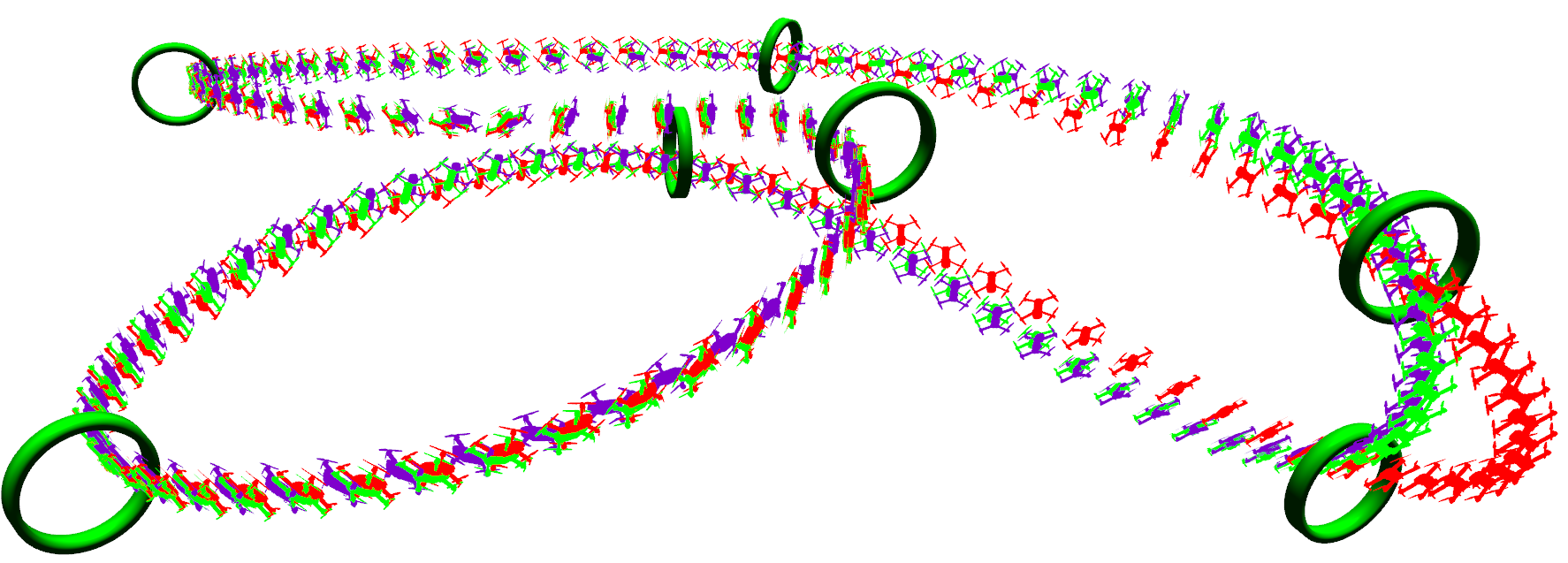}
    \caption{CPC and our raceline approaches for obstacle-free race scenario with narrow gates. CPC is shown in purple and our global and parametric frame racelines are colored green and red respectively. Our global frame raceline overlaps that of CPC while centerline regularity constraints force the parametric raceline to make a wider turn on the right. Snapshots of drone pose are shown every 50$\unit{\milli\second}$.}
    \label{fig:r1_cpc}
\end{figure}

We first narrowed the gates of the race scenario to circles of radius 0.6$\unit{\meter}$ to validate our approaches on the results of \cite{scaramuzza_cpc}. This yielded a collision-free region of width 0.3$\unit{\meter}$, identical to the distance tolerance used in \cite{scaramuzza_cpc}. We solved all racelines and ran the source code of \cite{scaramuzza_cpc} on the same computer and dependencies to obtain unbiased timing statistics. Racelines with this setup are shown in Figure \ref{fig:r1_cpc} and timing statistics are reported in Table \ref{tab:r1-comparison_table}. 

Our approaches compute racelines 100x as fast as \cite{scaramuzza_cpc} even when including setup time and solving for a point mass warmstart. The lap time extracted from \cite{scaramuzza_cpc} is within several milliseconds of the global frame raceline while the parametric frame raceline is slower as regularity constraints \eqref{eq:regularity_limit} force a wider vertical u-turn for the chosen centerline. This is a limitation of the centerline itself, not the raceline approach.

\begin{table}
\caption{Race Scenario Results - Narrow Gates}
\begin{tabular}{c c c c c c}
\hline
   Times: (s) & Lap & IPOPT & feval & Total Solve & Setup\\
\hline
Global Drone          & 6.096 & 14.34 & 0.940 & 15.28 & 7.811\\
Parametric Drone      & 6.173 & 9.250 & 0.830 & 10.079 & 8.649\\
Global Point Mass     & 6.091 & 0.820 & 0.077 & 0.897 & 1.778\\
Param. Point Mass     & 6.173 & 1.204 & 0.108 & 1.312 & 2.190\\
CPC \cite{scaramuzza_cpc} & 6.100 & 1630 & 1363 & 2993 & --- \\
\hline
\end{tabular}
\label{tab:r1-comparison_table}
\end{table}

\begin{figure}
    \centering
    \includegraphics[width = 0.6\linewidth]{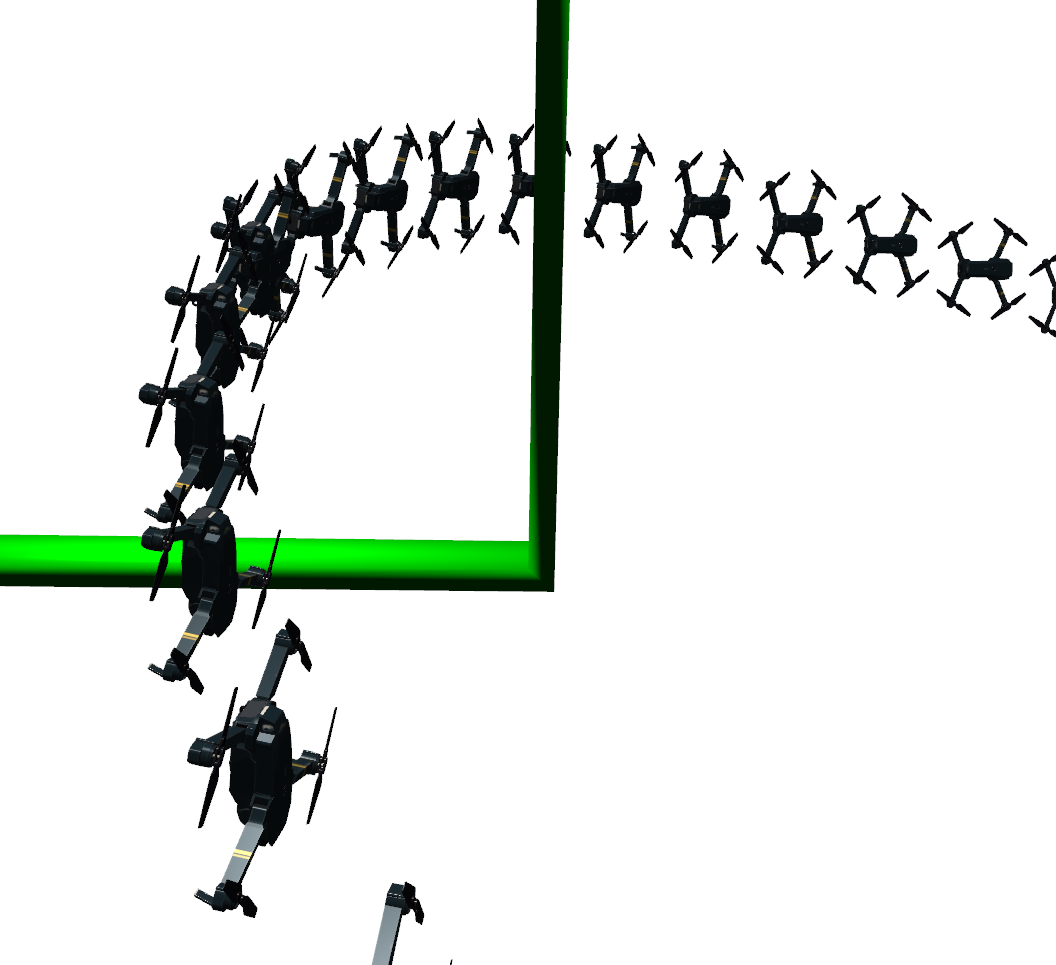}
    \caption{Closeup of \cite{scaramuzza_cpc} crossing finish line. Because the method of \cite{scaramuzza_cpc} solves several laps in one optimization problem, extracting a periodic raceline may be difficult. As shown, their method has discontinuous yaw angle whereas our methods solve directly for a periodic raceline.}
    \label{fig:r1_cpc_discontinuity}
\end{figure}

Unlike our approaches, CPC plans for roughly 2.5 laps of data with given initial and final states. A periodic raceline must be spliced from this data and drone orientation may not be continuous at the endpoints. This is shown in Figure \ref{fig:r1_cpc_discontinuity} where the yaw degree of freedom of the drone is discontinuous as a result of this splicing, and may have undesirable effects when the CPC trajectory is tracked. Our approaches directly solve for a periodic raceline.

\begin{figure}
    \centering
    \includegraphics[width=\linewidth]{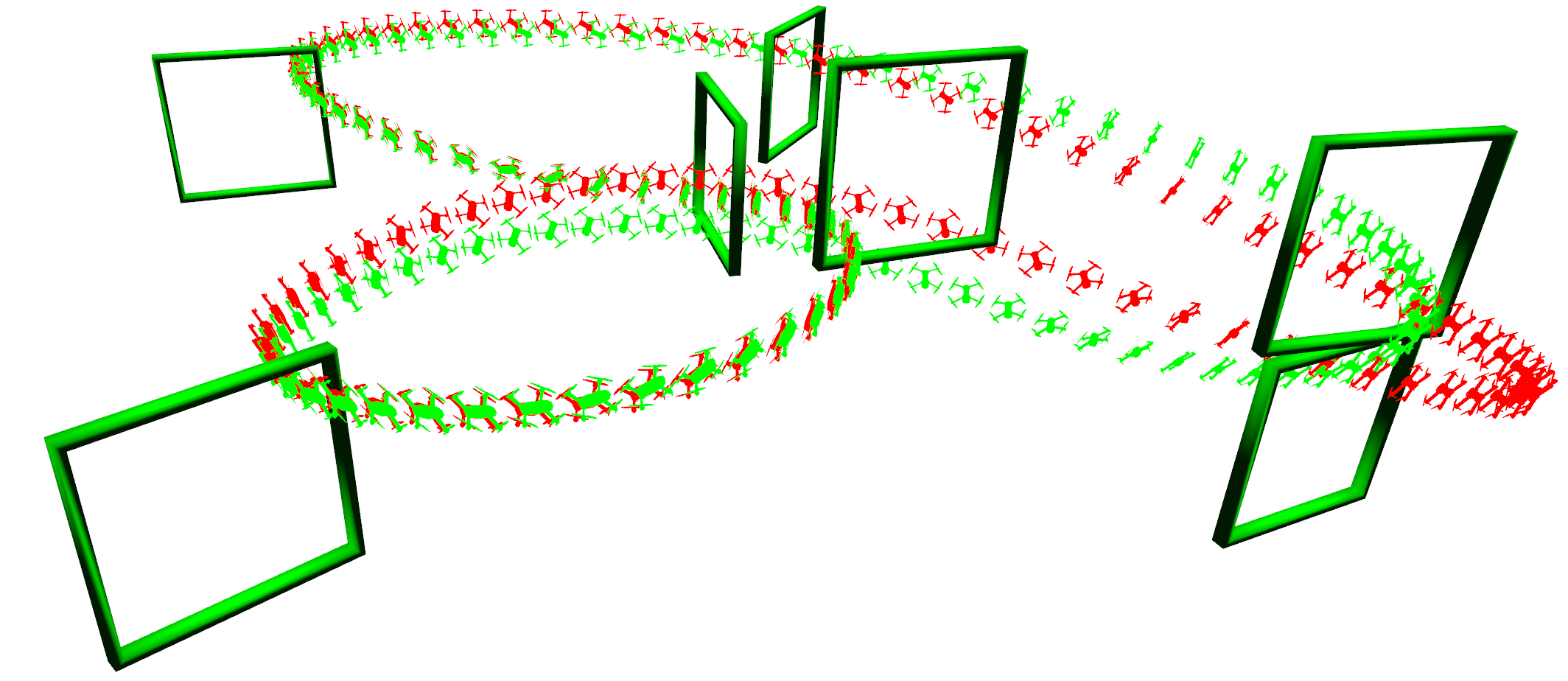}
    \caption{Global (green) and parametric (red) drone racelines for obstacle-free race scenario with full-size gates. The global raceline cuts the turn at far right tighter than the parametric raceline due to regularity constraints on the latter, yet both achieve tigher turns elsewhere than CPC, and a faster overall lap time. Snapshots of drone pose are shown every 50$\unit{\milli\second}$. An animation of the global frame trajectory is provided at \href{https://github.com/thomasfork/aircraft_trajectory_optimization}{github.com/thomasfork/aircraft\_trajectory\_optimization}.}
    \label{fig:r1_racelines}
\end{figure}

\begin{table}
\caption{Race Scenario Results - Full Size Gates}
\begin{tabular}{c c c c c c}
\hline
   Times: (s) & Lap & IPOPT & feval & Total Solve & Setup\\
\hline
Global Drone          & 5.648 & 7.442 & 0.552 & 7.994 & 7.200 \\
Parametric Drone      & 5.794 & 10.568 & 0.973 &  11.542 & 8.590 \\
Global Point Mass     & 5.645 & 0.859 & 0.082 & 0.941 & 1.612 \\
Param. Point Mass     & 5.788 & 1.158 & 0.111 & 1.269 & 2.039 \\
\hline
\end{tabular}
\label{tab:r1-table}
\end{table}

\begin{table}
\caption{Race Scenario Results - Full Gates and RK4}
\begin{tabular}{c c c c c c}
\hline
   Times: (s) & Lap & IPOPT & feval & Total Solve & Setup\\
\hline
Global Drone          & 5.651 & 13.765 & 18.749 & 32.513 & 10.668 \\
Parametric Drone      & 5.813 & 6.198  & 13.864 & 20.062 & 16.031 \\
Global Point Mass     & 5.646 & 0.599 & 0.023 & 0.622 & 0.707 \\
Param. Point Mass     & 5.813 & 1.226 & 0.321 & 1.547 & 1.835 \\
\hline
\end{tabular}
\label{tab:r1-table_rk4}
\end{table}

Whereas \cite{scaramuzza_cpc} requires a vehicle to pass through waypoints, our approaches may leverage the full area of convex gates for trajectory optimization since we control exactly where the trajectory passes through gates. To illustrate this, we imposed constraints on $y$ and $n$ that spanned the entire collision-free area of the gates. This allowed for tighter turns shown in Figure \ref{fig:r1_racelines} and resulted in much faster lap times reported in Table \ref{tab:r1-table}. Once more, the parametric raceline was forced to make a wider turn on the vertical u-turn due to regularity constraints \eqref{eq:regularity_limit}, yet achieved a faster lap than the CPC raceline by making much tighter turns everywhere else.

As a secondary comparison, we implemented the same problem with direct multiple shooting using 4\textsuperscript{th} order Runge-Kutta integration instead of collocation. We increased the number of finite elements per gate to 70 to keep the spacing of decision variables consistent. Timing statistics are reported in Table \ref{tab:r1-table_rk4}. Overall the racelines are identical but compute time is slower, largely due to the function evaluation complexity of the integration scheme, seen in the ``feval" column.


\subsection{Figure-Eight Scenario} \label{sec:r2}
\begin{figure}
    \centering
    \includegraphics[width = 0.50\linewidth]{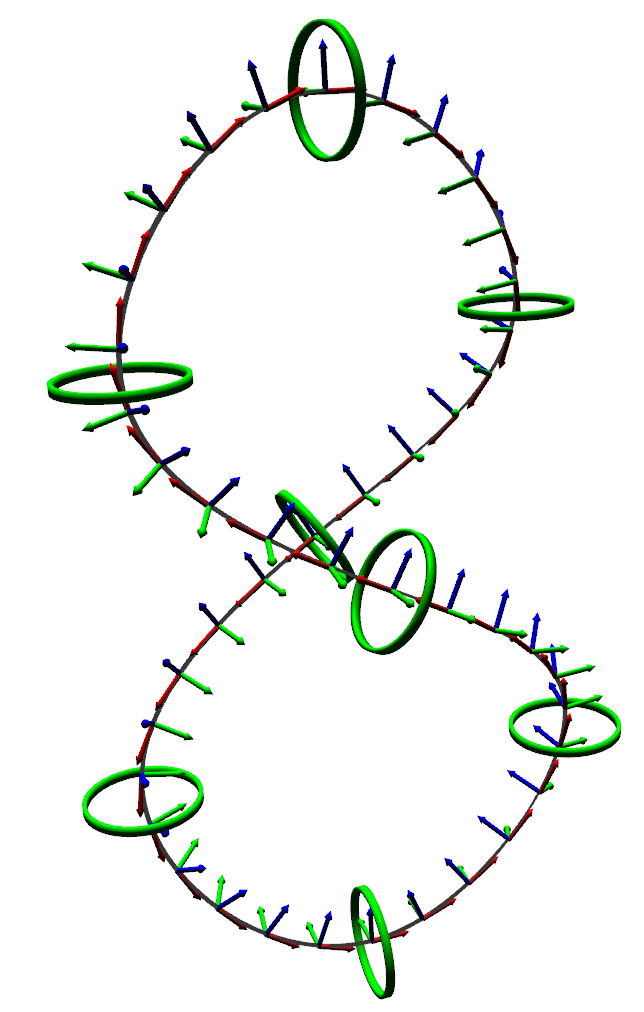}
    \caption{Figure-eight scenario for Section \ref{sec:r2}. Centerline and parametric frame are shown with the same color scheme as Figure \ref{fig:r1_scenario}. The diameter of each end of the figure-eight is 10$\unit{\meter}$ and they cross 4$\unit{\meter}$ apart. Importantly, the time-optimal trajectory through the upper loop involves orienting the drone upside-down and the drone must rotate 180 degrees between the upper and lower loops, exacerbating nonconvexity and resulting in convergence issues without a warmstart.}
    \label{fig:r2_scenario}
\end{figure}

We now demonstrate improved convergence on a more challenging scenario. We introduce the scenario shown in Figure \ref{fig:r2_scenario}: a vertical figure-eight, widened in the middle to avoid overlapping gates. Regularity limits from equation \eqref{eq:regularity_limit} do not impact the middle of the figure-eight as they only depend on the local shape of the centerline: its curvature. Global shape, such as the centerline crossing over itself does not impact \eqref{eq:regularity_limit}, but can complicate associating known global frame position to non-Euclidean coordinates.

The figure-eight scenario is such that the time-optimal drone trajectory through the upper portion involves rotating the drone upside-down. This means that without a warmstart, optimization algorithms start much further from optimal than in Section \ref{sec:r1}, exacerbating nonconvexity and convergence issues. Furthermore, any drone raceline computed with Euler angles will be suboptimal from having to avoid gimbal lock. However, switching to a quaternion orientation parameterization does not solve these issues, as nonconvexity effects persist, we demonstrate this shortly.

Unlike the previous scenario, regularity constraints \eqref{eq:regularity_limit} do not impact the final trajectory - global and parametric frame racelines were found to have identical lap times as gate constraint became active before \eqref{eq:regularity_limit}. We focus on our non-Euclidean raceline approach, for which we set up versions with Euler angle and quaternion rotation parameterizations to compare the two and illustrate issues with each. 50 collocation intervals were used for all of our approaches.

\begin{figure}
    \centering
    \includegraphics[width=.45\linewidth]{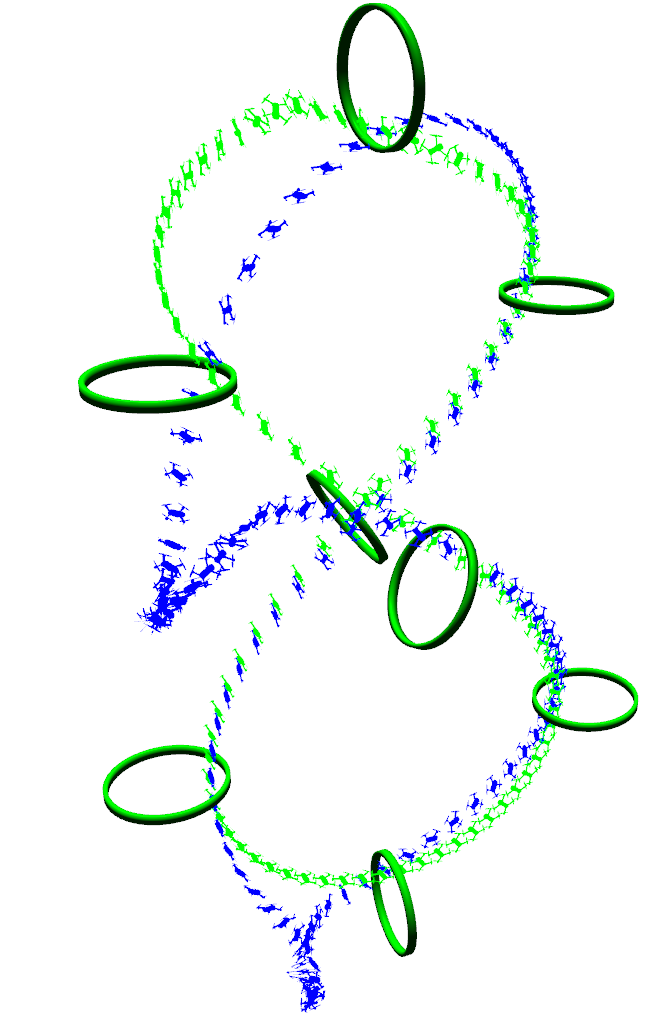}
    \caption{Racelines computed with our non-Euclidean approach on the figure-eight scenario with warmstart disabled. The blue raceline was computed with quaternion-based orientation and the green with Euler angles. With Euler angles the drone cannot pitch or roll enough for the upper loop as constraints are present to avoid gimbal lock. With a quaternion, spurious loops in the trajectory emerge in the trajectory. Snapshots of drone pose are shown every 50$\unit{\milli\second}$.}
    \label{fig:r2_no_ws}
\end{figure}

\begin{figure}
    \centering
    \includegraphics[width=.45\linewidth]{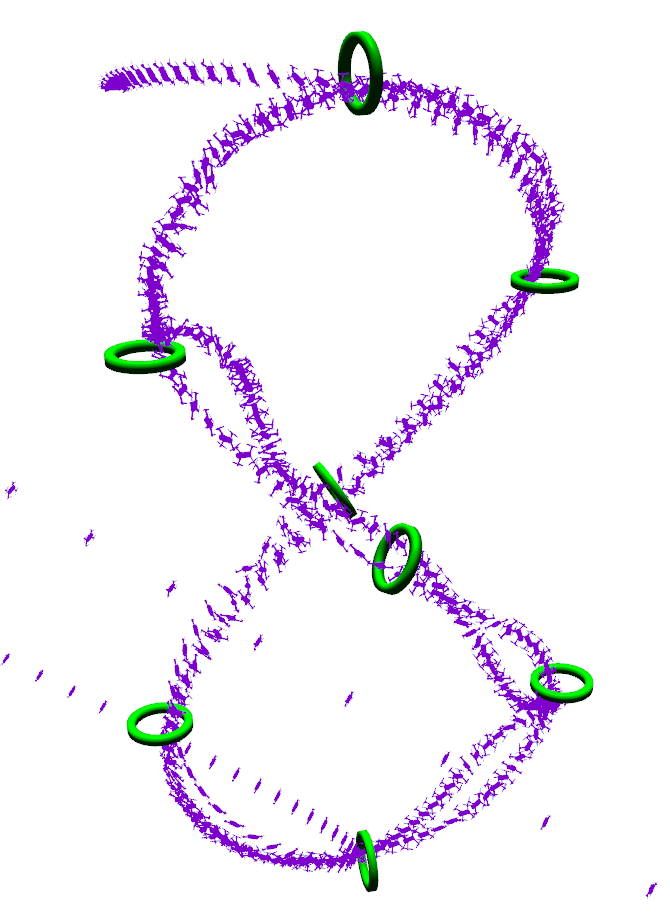}
    \caption{Raceline computed using the approach of \cite{scaramuzza_cpc} for the figure-eight scenario. The full trajectory is shown as the algorithm did not converge. Gates are shrunk to visualize the distance tolerance of the algorithm. Although the shape of one lap is nearly correct, the drone spins erratically throughout, and finally diverges 700$\unit{\meter}$ at the end. Snapshots of drone pose are shown every 50$\unit{\milli\second}$.}
    \label{fig:r2_cpc}
\end{figure}

We computed racelines with CPC and with our non-Euclidean approach without a warmstart to illustrate the difficulty of the figure-eight scenario for trajectory optimization. Racelines for our non-Euclidean approach and CPC are shown in Figures \ref{fig:r2_no_ws} and \ref{fig:r2_cpc} respectively. With our non-Euclidean approach we show racelines for both Euler angles and quaternion orientation parameterizations. 

Our non-Euclidean approach converges when using Euler angles, however the drone is unable to rotate upside-down due to gimbal lock limits and the resulting raceline is evidently suboptimal. Meanwhile with a quaternion, although the algorithm converges, there are two unecessary loops in the trajectory, and the overall trajectory is slower than that computed with Euler angles. Evidently, simply avoiding the use of Euler angles is insufficient. 

On the same scenario, CPC failed to converge after 4 hours and 10000 iterations in IPOPT. We show the full result of the last iteration in Figure \ref{fig:r2_cpc} as the algorithm failed to converge to a periodic raceline. Although one of the two laps has appropriate trajectory shape, it involves the drone spinning erratically and unpredictably with greatly reduced speed. The end of the CPC trajectory then travels 700$\unit{\meter}$ away from the scenario despite constraints on initial and final position, illustrating failure of the algorithm to converge.

\begin{figure}
    \centering
    \includegraphics[width = 0.45\linewidth]{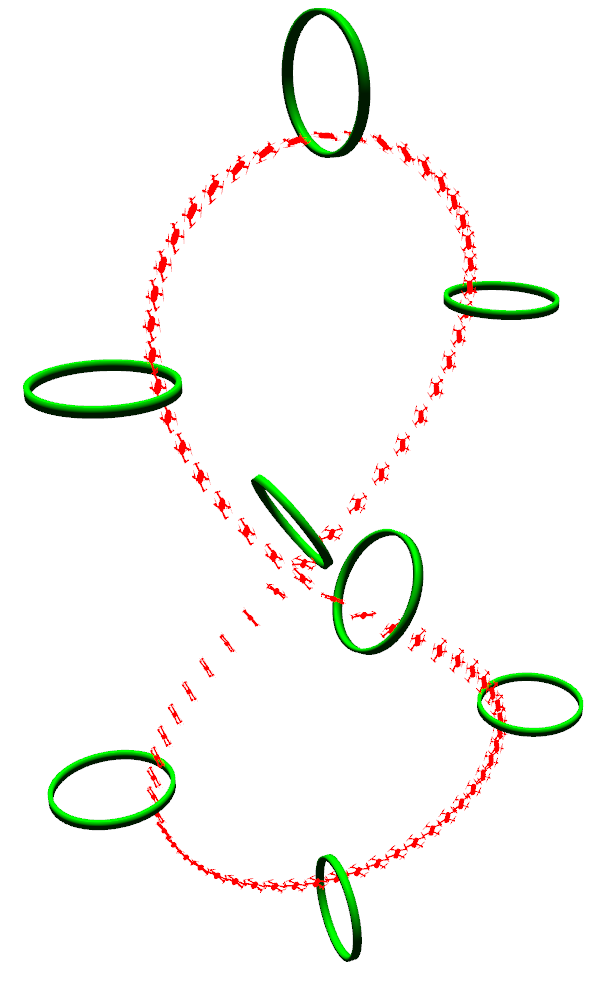}
    \caption{Drone raceline computed by our non-Euclidean approach with a point mass raceline used to warmstart pose and velocity. Drone orientation is expressed by a quaternion during optimization, avoiding gimbal lock, and the initial orientation warmstart derived from the thrust of the point-mass raceline is sufficient for convergence. Snapshots of drone pose are shown every 50$\unit{\milli\second}$. An animation is provided at \href{https://github.com/thomasfork/aircraft_trajectory_optimization}{github.com/thomasfork/aircraft\_trajectory\_optimization}.}
    \label{fig:r2_raceline}
\end{figure}

To illustrate the importance of warmstarting, using a quaternion and the robustness of the overall approach, we then ran our non-Euclidean approach with quaternion orientation using the point mass warmstart procedure. With this the algorithm converged issue-free to a faster lap than other test cases (Figure \ref{fig:r2_raceline}). Similar warmstarting for the CPC solver was not investigated in \cite{scaramuzza_cpc} and was not implemented here. Timing statistics for all approaches are reported in Table \ref{tab:r2-table}.

\begin{table}
\caption{Figure-Eight Scenario Results}
\begin{tabular}{c c c c c c}
\hline
   Times: (s) & Lap & IPOPT & feval & Total Solve & Setup\\
\hline
Warmstart (quat)     & 4.294 & 15.65 & 0.986 & 16.64 & 8.357\\
Coldstart (quat) & 6.165 & 111.7 & 7.30 & 119.0 & 6.396\\
Coldstart (Euler)      & 4.674 & 21.075 & 1.553 & 22.63 & 5.418\\
Point Mass             & 4.289 & 0.986 & 0.093 & 1.079 & 2.323\\
CPC \cite{scaramuzza_cpc} & ---   & 8780  & 5972  & 14752 & --- \\
\hline
\multicolumn{6}{r}{(Racelines in parametric frame)} \\
\end{tabular}
\label{tab:r2-table}
\end{table}

\subsection{Obstacle Avoidance Scenario} \label{sec:r3}
\begin{figure}
    \centering
    \includegraphics[width = \linewidth]{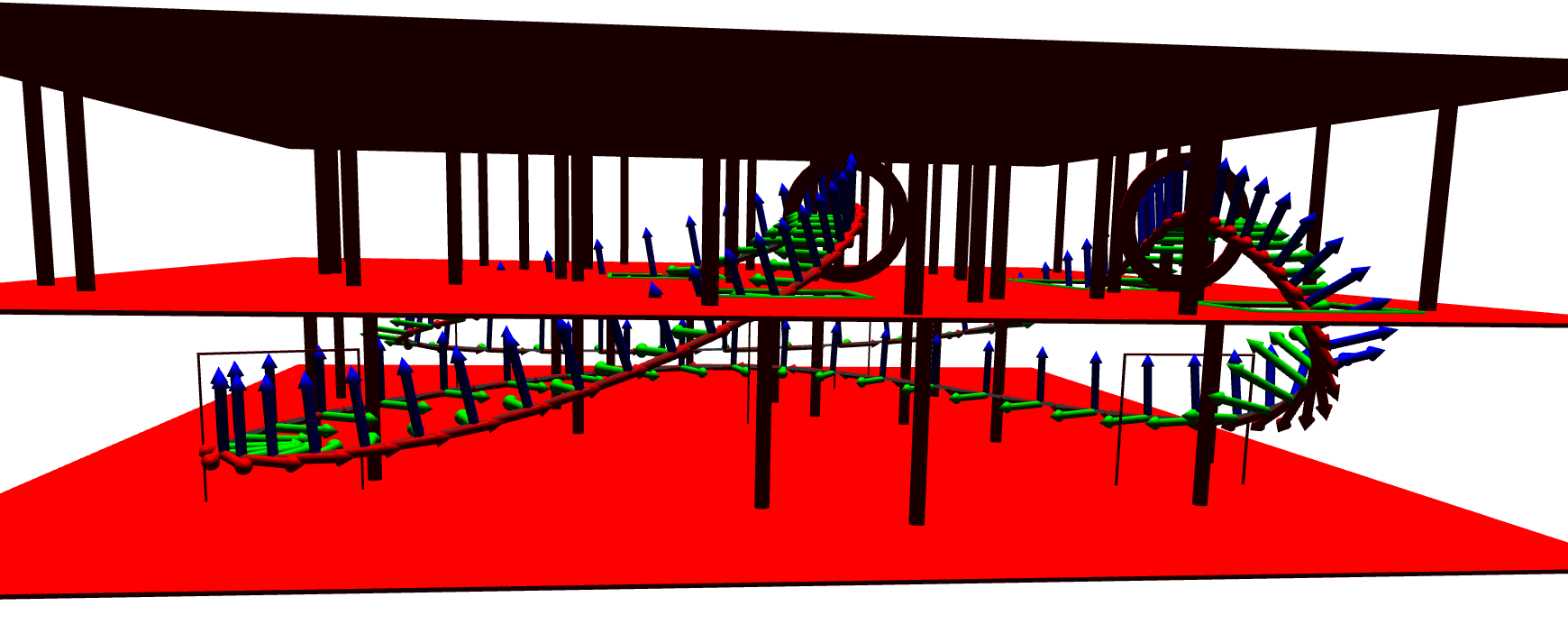}
    \caption{Obstacle avoidance scenario, adapted from \cite{scaramuzza_waypoint_search}. Centerline and parametric frame coordinate systems are shown with color scheme of previous figures \ref{fig:r1_scenario} and \ref{fig:r2_scenario}. The obstacle-rich environment is shown in red, with pillars darkened. The goal is to compute a periodic raceline that avoids all pillars while passing through a sequence of gates, which are also defined as obstacles.}
    \label{fig:r3_scenario}
\end{figure}

Finally, we applied our non-Euclidean approach to the obstacle-rich environment of \cite{scaramuzza_waypoint_search}. This environment is similar to the one in Section \ref{sec:r1} with the addition of many obstacles and a slightly different gate layout, shown in Figure \ref{fig:r3_scenario}. We did not compare our method to CPC or \cite{scaramuzza_waypoint_search} as \cite{scaramuzza_waypoint_search} focused on search-based approaches for initial collision-free trajectories rather than racelines and reported that CPC did not converge for this scenario.

\begin{figure}
    \centering
    \includegraphics[width = \linewidth]{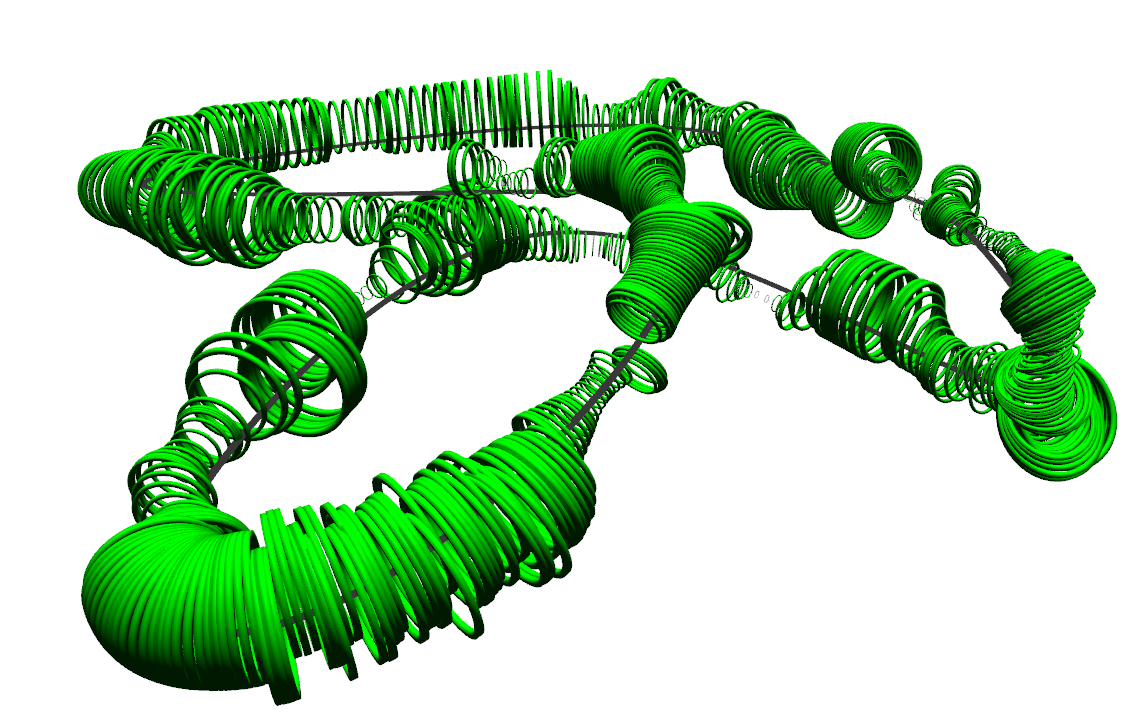}
    \caption{Centerline and safe flight corridor for a collision avoidance radius of $0.4\unit{\meter}$. Rings illustrating feasible position are shown for every decision variable and enforced as constraints during optimization. Nonuniform spacing is a consequence of the collocation method used. At times the centerline is not part of the safe flight corridor, indicating that the centerline was infeasible for this radius.}
    \label{fig:r3_tube}
\end{figure}

We computed a safe flight corridor using the approach of Section \ref{sec:obstacle_avoidance}. This is shown in Figure \ref{fig:r3_tube}, where each ring corresponds to a discretization point along the centerline and the interior of each ring is the set of feasible positions for the drone center of mass at this point. These rings correspond to points where the signed distance of the drone to any obstacle exceeds a specified collision avoidance distance, meaning that 3D obstacle avoidance is locally satisfied with 2D constraints.

\begin{figure}
    \centering
    \includegraphics[width = \linewidth]{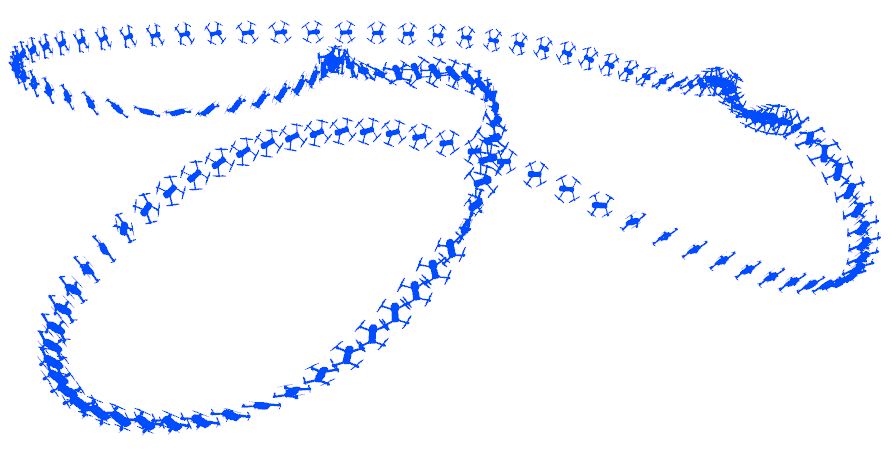}
    \caption{Complete drone raceline computed with collision avoidance radius of $0.4\unit{\meter}$. Obstacles have been hidden to view the entire raceline but force several complex maneuvers between pillars and gates. A detailed closeup of the middle portion is shown in Figure \ref{fig:r3_closeup}. Snapshots of drone pose are shown every 50$\unit{\milli\second}$. An animation is provided at \href{https://github.com/thomasfork/aircraft_trajectory_optimization}{github.com/thomasfork/aircraft\_trajectory\_optimization}.}
    \label{fig:r3_raceline}
\end{figure}

\begin{figure}
    \centering
    \includegraphics[width = \linewidth]{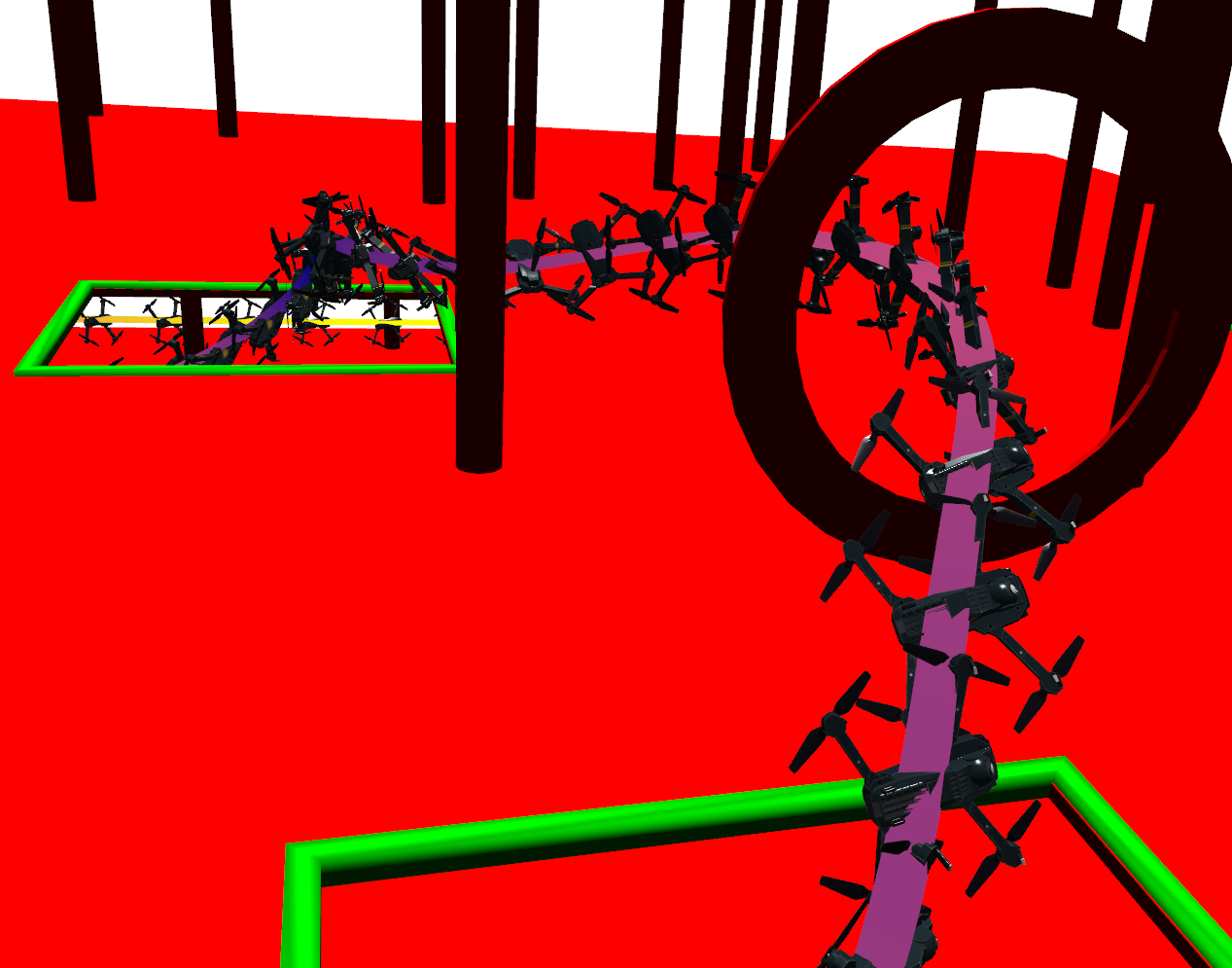}
    \caption{Closeup of computed drone raceline with collision avoidance radius of $0.4\unit{\meter}$. The optimal raceline undergoes large changes in drone orientation necessary to swerve through gates and avoid obstacles at high speed. Snapshots are shown every 50$\unit{\milli\second}$.}
    \label{fig:r3_closeup}
\end{figure}

Safe flight corridor and accompanying racelines were computed with collision avoidance radii of 0.2$\unit{\meter}$, 0.3$\unit{\meter}$ and 0.4$\unit{\meter}$. The 0.4$\unit{\meter}$ SFC is shown in Figure \ref{fig:r3_tube}, with the other two corresponding to wider rings in identical position. We also computed racelines with collision avoidance constraints disabled and replaced with gates at each of the waypoints used in \cite{scaramuzza_waypoint_search} (as in Section \ref{sec:r1} and \ref{sec:r2}). All racelines used 100 collocation intervals, evenly spread as seen by the spacing of the rings in Figure \ref{fig:r3_tube}. We show the drone raceline for 0.4$\unit{\meter}$ in Figures \ref{fig:r3_raceline} and \ref{fig:r3_closeup}. Timing statistics for all four cases are reported in Table \ref{tab:r3-table}. 

\begin{table}
\caption{Obstacle Avoidance Scenario Results}
\begin{tabular}{c c c c c c}
\hline
   Times: (s) & Lap & IPOPT & feval & Total Solve & Setup\\
\hline
\multicolumn{3}{l}{Avoidance radius of $0.4\unit{\meter}$} \\
Drone      & 7.435 & 20.911 & 1.715 & 22.626 & 16.820\\
Point Mass & 7.386 & 7.560 & 0.474 & 8.034 & 3.049\\
\hline
\multicolumn{3}{l}{Avoidance radius of $0.3\unit{\meter}$} \\
Drone      & 6.196 & 36.824 & 2.565 & 39.388 & 15.398\\
Point Mass & 6.195 & 4.274 & 0.278 & 4.552 & 3.552\\
\hline
\multicolumn{3}{l}{Avoidance radius of $0.2\unit{\meter}$} \\
Drone      & 6.055 & 34.594 & 2.252 & 36.847 & 12.190\\
Point Mass & 6.091 & 2.452 & 0.173 & 2.626 & 3.011\\
\hline
\multicolumn{3}{l}{No Obstacle Avoidance} \\
Drone      & 5.577 & 15.629 & 1.063 & 16.692 & 13.195\\
Point Mass & 5.580 & 3.068 & 0.213 & 3.281 & 3.674\\

\end{tabular}
\label{tab:r3-table}
\end{table}

\section{Conclusion}\label{sec:conclusion}

We presented two approaches for trajectory optimization of aircraft based on Euclidean geometry and a novel non-Euclidean approach. We demonstrated 100x faster compute time for both approaches compared to previous methods, as well as improved reliability and lap time. We presented a novel method for obstacle avoidance using the non-Euclidean approach and demonstrated its use to compute high fidelity racelines in an obstacle-rich environment.

\section{Acknowledgements}

Source code used for comparison with \cite{scaramuzza_cpc} was downloaded from \href{https://zenodo.org/record/5036287}{zenodo.org/record/5036287}

The obstacle-rich environment of \cite{scaramuzza_waypoint_search} was downloaded from \href{https://github.com/uzh-rpg/sb_min_time_quadrotor_planning}{github.com/uzh-rpg/sb\_min\_time\_quadrotor\_planning}

The graphical quadrotor model is \href{https://creativecommons.org/licenses/by/4.0/}{CC BY 4.0}, made by Sketchfab user \texttt{the\_Thorminator}
  \href{https://skfb.ly/orIx8}{skfb.ly/orIx8} and was recolored for several figures.

\renewcommand*{\bibfont}{\footnotesize}
\printbibliography

@book{mechanics_landau_lifshitz,
    Author = {L. D. Landau and E. M. Lifshitz},
    Title = {Mechanics},
    subtitle = {Course of Theoretical Physics},
    publisher = {Butterworth-Heinemann},
    Year = {1976},
    Month = {1},
%    address = {Oxford, U.K.},
    Volume = {1},
    Edition = {3}
}

@book{differential_geometry_of_curves_and_surfaces,
    author = {Manfredo P. do Carmen},
    title = {Differential geometry of curves and surfaces},
    publisher = {Prentice Hall},
    year = {1976}
}

@book{murray_robotics,
    author = {Richard M. Murray and Zexiang Li and S. Shankar Sastry},
    title = {A Mathematical Introduction to Robotic Manipulation},
    year = {1994},
    publisher = {CRC Press,}}

@book{intro_relativity,
    title = {Introduction to General Relativity},
    author = {Ronald Adler and Maurice Bazin and Menahem Shiffer},
    publisher = {McGraw-Hill Book Company},
    year = {1965}
}

@article{gpops_ii,
author = {Patterson, Michael A. and Rao, Anil V.},
title = {GPOPS-II: A MATLAB Software for Solving Multiple-Phase Optimal Control Problems Using Hp-Adaptive Gaussian Quadrature Collocation Methods and Sparse Nonlinear Programming},
year = {2014},
issue_date = {October 2014},
publisher = {Association for Computing Machinery},
address = {New York, NY, USA},
volume = {41},
number = {1},
journal = {ACM Trans. Math. Softw.},
month = {10},
articleno = {1},
numpages = {37}
}

@book{limebeer_car_racing_survey,
    title = {Dynamics and optimal control of road vehicles},
    author = {D. J. N. Limebeer and M. Massaro},
    publisher = {Oxford University Press},
    year = {2018}}

@article{3d_part_1,
    author = {Perantoni, Giacomo and Limebeer, David J. N.},
    title = "{Optimal Control of a Formula One Car on a Three-Dimensional Track—Part 1: Track Modeling and Identification}",
    journal = {Journal of Dynamic Systems, Measurement, and Control},
    volume = {137},
    number = {5},
    year = {2015},
    month = {05}
}

@ARTICLE{obca,
  author={Zhang, Xiaojing and Liniger, Alexander and Borrelli, Francesco},
  journal={IEEE Transactions on Control Systems Technology}, 
  title={Optimization-Based Collision Avoidance}, 
  year={2021},
  volume={29},
  number={3},
  pages={972-983}}

@INPROCEEDINGS{planning_convex_decomp,
  author={Deits, Robin and Tedrake, Russ},
  booktitle={2015 IEEE International Conference on Robotics and Automation (ICRA)}, 
  title={Efficient mixed-integer planning for UAVs in cluttered environments}, 
  year={2015},
  volume={},
  number={},
  pages={42-49}}

@INPROCEEDINGS{miqp_polyhedra_following,
  author={Tordesillas, Jesus and Lopez, Brett T. and How, Jonathan P.},
  booktitle={2019 IEEE/RSJ International Conference on Intelligent Robots and Systems (IROS)}, 
  title={FASTER: Fast and Safe Trajectory Planner for Flights in Unknown Environments}, 
  year={2019},
  volume={},
  number={},
  pages={1934-1940}}

@ARTICLE{flatness_trajectory_opt,
  author={Wang, Zhepei and Zhou, Xin and Xu, Chao and Gao, Fei},
  journal={IEEE Transactions on Robotics}, 
  title={Geometrically Constrained Trajectory Optimization for Multicopters}, 
  year={2022},
  volume={38},
  number={5},
  pages={3259-3278}}

@article{scaramuzza_mpcc,
  author={Romero, Angel and Sun, Sihao and Foehn, Philipp and Scaramuzza, Davide},
  journal={IEEE Transactions on Robotics}, 
  title={Model Predictive Contouring Control for Time-Optimal Quadrotor Flight}, 
  year={2022},
  volume={38},
  number={6},
  pages={3340-3356}}

@inproceedings{scaramuzza_deep_acrobatics,
    title = {Deep Drone Acrobatics},
    author = {Elia Kaufmann and Antonio Loquercio and René Ranftl and Matthias Müller and Vladlen Koltun and Davide Scaramuzza },
    booktitle = {Robotics: Science and Systems},
    year = {2020},
    pages = {},
    volume = {},
    number = {}}

@article{scaramuzza_cpc,
author = {Philipp Foehn  and Angel Romero  and Davide Scaramuzza },
title = {Time-optimal planning for quadrotor waypoint flight},
journal = {Science Robotics},
volume = {6},
number = {56},
pages = {eabh1221},
year = {2021}}

@misc{scaramuzza_racing_survey,
  url = {https://arxiv.org/abs/2301.01755},
  author = {Hanover, Drew and Loquercio, Antonio and Bauersfeld, Leonard and Romero, Angel and Penicka, Robert and Song, Yunlong and Cioffi, Giovanni and Kaufmann, Elia and Scaramuzza, Davide},
  title = {Autonomous Drone Racing: A Survey},
  publisher = {arXiv},
  year = {2023},
  copyright = {arXiv.org perpetual, non-exclusive license}
}

@ARTICLE{scaramuzza_waypoint_search,
  author={Penicka, Robert and Scaramuzza, Davide},
  journal={IEEE Robotics and Automation Letters}, 
  title={Minimum-Time Quadrotor Waypoint Flight in Cluttered Environments}, 
  year={2022},
  volume={7},
  number={2},
  pages={5719-5726}}

@article{drone_search_planning,
author = {Liu, Sikang and Mohta, Kartik and Atanasov, Nikolay and Kondepogu, Vijayakumar},
year = {2017},
month = {10},
pages = {},
title = {Search-Based Motion Planning for Aggressive Flight in SE(3)},
volume = {PP},
journal = {IEEE Robotics and Automation Letters}
}

@ARTICLE{drone_navigation,
  author={Jung, Sunggoo and Hwang, Sunyou and Shin, Heemin and Shim, David Hyunchul},
  journal={IEEE Robotics and Automation Letters}, 
  title={Perception, Guidance, and Navigation for Indoor Autonomous Drone Racing Using Deep Learning}, 
  year={2018},
  volume={3},
  number={3},
  pages={2539-2544}}

@inproceedings{arrizabalaga2022tunnels,
  title={Towards time-optimal tunnel-following for quadrotors},
  author={Arrizabalaga, Jon and Ryll, Markus},
  booktitle={2022 International Conference on Robotics and Automation (ICRA)},
  pages={4044--4050},
  year={2022},
  organization={IEEE}
}

@article{arrizabalaga2022spatially,
  title={Spatially Constrained Time-Optimal Motion Planning},
  author={Arrizabalaga, Jon and Ryll, Markus},
  journal={arXiv preprint arXiv:2210.02345},
  year={2022}
}

@inproceedings{airplane_racing,
    author = {F. Fisch and J. Lenz and F. Holzapfel and G. Sachs},
    title={Trajectory Optimization Applied to Air Races},
    year = {2009},
    day = {10},
    month = {8},
    booktitle={AIAA Atmospheric Flight Mechanics Conference}}

@ARTICLE{giuseppe_drone_racing,
  author={Spedicato, Sara and Notarstefano, Giuseppe},
  journal={IEEE Transactions on Control Systems Technology}, 
  title={Minimum-Time Trajectory Generation for Quadrotors in Constrained Environments}, 
  year={2018},
  volume={26},
  number={4},
  pages={1335-1344}}

@INPROCEEDINGS{drone_min_snap,
  author={Mellinger, Daniel and Kumar, Vijay},
  booktitle={2011 IEEE International Conference on Robotics and Automation}, 
  title={Minimum snap trajectory generation and control for quadrotors}, 
  year={2011},
  volume={},
  number={},
  pages={2520-2525}}

@inproceedings{red_bull_racing,
    author = {M. Bittner and F. Fisch and J. Lenz and F. Holzapfel},
    title={A Multi-Model Gauss Pseudospectral Optimization Method for Aircraft Trajectories},
    year = {2012},
    day = {13},
    month = {8},
    booktitle={AIAA Atmospheric Flight Mechanics Conference}}

@INPROCEEDINGS{mueller_drone_traj_opt,
  author={Mueller, Mark W. and Hehn, Markus and D'Andrea, Raffaello},
  booktitle={2013 IEEE/RSJ International Conference on Intelligent Robots and Systems}, 
  title={A computationally efficient algorithm for state-to-state quadrocopter trajectory generation and feasibility verification}, 
  year={2013},
  volume={},
  number={},
  pages={3480-3486}}

@ARTICLE{safe_flight_corridors,
  author={Liu, Sikang and Watterson, Michael and Mohta, Kartik and Sun, Ke and Bhattacharya, Subhrajit and Taylor, Camillo J. and Kumar, Vijay},
  journal={IEEE Robotics and Automation Letters}, 
  title={Planning Dynamically Feasible Trajectories for Quadrotors Using Safe Flight Corridors in 3-D Complex Environments}, 
  year={2017},
  volume={2},
  number={3},
  pages={1688-1695}}

@article{aggarwal_path_planning_survey,
  title={Path planning techniques for unmanned aerial vehicles: A review, solutions, and challenges},
  author={Aggarwal, Shubhani and Kumar, Neeraj},
  journal={Computer Communications},
  volume={149},
  pages={270--299},
  year={2020},
  publisher={Elsevier}
}

@inproceedings{richards2002aircraft,
  title={Aircraft trajectory planning with collision avoidance using mixed integer linear programming},
  author={Richards, Arthur and How, Jonathan P},
  booktitle={Proceedings of the 2002 American Control Conference (IEEE Cat. No. CH37301)},
  volume={3},
  pages={1936--1941},
  year={2002},
  organization={IEEE}
}

@Article{Andersson2018,
  Author = {Joel A E Andersson and Joris Gillis and Greg Horn
            and James B Rawlings and Moritz Diehl},
  Title = {{CasADi} -- {A} software framework for nonlinear optimization
           and optimal control},
  Journal = {Mathematical Programming Computation},
  Year = {2018},
}

@article{ipopt,
author = {Wächter, Andreas and Biegler, Lorenz},
year = {2006},
month = {03},
pages = {25-57},
title = {On the Implementation of an Interior-Point Filter Line-Search Algorithm for Large-Scale Nonlinear Programming},
volume = {106},
journal = {Mathematical programming}
}

@article{MUMPS_1,
   title   = {A Fully Asynchronous Multifrontal Solver Using Distributed Dynamic Scheduling},
   author  = {P.R. Amestoy and I. S. Duff and J. Koster and J.-Y. L'Excellent},
   journal = {SIAM Journal on Matrix Analysis and Applications},
   volume  = {23},
   number  = {1},
   year    = {2001},
   pages   = {15-41}
 }

@article{zhou2023efficient,
  title={Efficient and robust time-optimal trajectory planning and control for agile quadrotor flight},
  author={Zhou, Ziyu and Wang, Gang and Sun, Jian and Wang, Jikai and Chen, Jie},
  journal={IEEE Robotics and Automation Letters},
  year={2023},
  publisher={IEEE}
}

@article{ramirez2021gravity,
  title={A gravity-referenced moving frame for vehicle path following applications in 3d},
  author={Ramírez, Juan Carlos Hernández and Nahon, Meyer},
  journal={IEEE Robotics and Automation Letters},
  volume={6},
  number={3},
  pages={4393--4400},
  year={2021},
  publisher={IEEE}
}

@article{kelly2017introduction,
  title={An introduction to trajectory optimization: How to do your own direct collocation},
  author={Kelly, Matthew},
  journal={SIAM Review},
  volume={59},
  number={4},
  pages={849--904},
  year={2017},
  publisher={SIAM}
}

@article{vanroye2023fatrop,
  title={FATROP: A Fast Constrained Optimal Control Problem Solver for Robot Trajectory Optimization and Control},
  author={Vanroye, Lander and Sathya, Ajay and De Schutter, Joris and Decr{\'e}, Wilm},
  journal={arXiv preprint arXiv:2303.16746},
  year={2023}
}

@article{qin2023time,
  title={Time-Optimal Gate-Traversing Planner for Autonomous Drone Racing},
  author={Qin, Chao and Michet, Maxime SJ and Chen, Jingxiang and Liu, Hugh H-T},
  journal={arXiv preprint arXiv:2309.06837},
  year={2023}
}

@inproceedings{bos2022multi,
  title={Multi-stage optimal control problem formulation for drone racing through gates and tunnels},
  author={Bos, Mathias and Decr{\'e}, Wilm and Swevers, Jan and Pipeleers, Goele},
  booktitle={2022 IEEE 17th International Conference on Advanced Motion Control (AMC)},
  pages={376--382},
  year={2022},
  organization={IEEE}
}

@inproceedings{ramos2021minimum,
  title={Minimum-time optimal control for a quadcopter trajectory through gates with arbitrary pose},
  author={Ramos, Oscar E},
  booktitle={2021 IEEE XXVIII International Conference on Electronics, Electrical Engineering and Computing (INTERCON)},
  pages={1--4},
  year={2021},
  organization={IEEE}
}

@article{marcucci2024shortest,
  title={Shortest paths in graphs of convex sets},
  author={Marcucci, Tobia and Umenberger, Jack and Parrilo, Pablo and Tedrake, Russ},
  journal={SIAM Journal on Optimization},
  volume={34},
  number={1},
  pages={507--532},
  year={2024},
  publisher={SIAM}
}

@article{marcucci2023fast,
  title={Fast Path Planning Through Large Collections of Safe Boxes},
  author={Marcucci, Tobia and Nobel, Parth and Tedrake, Russ and Boyd, Stephen},
  journal={arXiv preprint arXiv:2305.01072},
  year={2023}
}
\appendix

\subsection{Vehicle Model Parameters} \label{app:parameters}

\begin{table}[h]
\begin{tabular}{c c c c}
\hline
$m$ & 1.0 \unit{\kilo\gram}                                 & $l$ & 0.15 \unit{\meter} \\
$g$ & 9.81 \unit{\kilo\gram\per\square\second}              & $c_\tau$ & 0.05 \unit{\meter} \\
$I^b_1$ & \num{1.0e-3} \unit{\kilo\gram\square\meter}       & TWR & 3.3\\
$I^b_2$ & \num{1.0e-3} \unit{\kilo\gram\square\meter}       & $T_{min}$ & 0.2 \unit{\newton}\\
$I^b_3$ & \num{1.7e-3} \unit{\kilo\gram\square\meter}\\
\hline
\end{tabular}
\end{table}
\subsection{Remarks on {\cite{giuseppe_drone_racing}}} \label{app:giuseppe}
In \cite{giuseppe_drone_racing}, a curvilinear coordinate is introduced as arc length along a curve. They use the direction of curvature of the curve as a lateral offset direction, with a second offset chosen orthonormal to this and the tangent of the curve. They use Frenet-Serret formulas with the resulting coordinate system, which assumes it is torsion-free \cite{differential_geometry_of_curves_and_surfaces}. However, there is no guarantee that the direction of curvature is torsion-free, continuous (eg. an ``S" bend) or even exists, such as for a straight line, meaning a correct differential equation for vehicle kinematics does not always exist with their approach.
\subsection{Remarks on {\cite{arrizabalaga2022spatially}}} \label{app:arrizabalaga}
In Section \ref{sec:kinematic_remarks} we introduced the regularity constraint \eqref{eq:regularity_limit}. \cite{arrizabalaga2022spatially} ignored this by discretizing with respect to $s$ and claiming $\partial_s z = \sfrac{\partial_t z}{\dot{s}}$, where $z$ is the state of a system and includes 3D position, seemingly avoiding the need for a regularity constraint. 
However, \cite{arrizabalaga2022spatially} ignored the fact that 3D position must be constrained to the $y-n$ plane at any given $s$. Enforcing this results in a vehicle being forced to stop or backtrack when at or past the limits of \eqref{eq:regularity_limit}. Not enforcing it means the resulting trajectory violates the original dynamics as consistency between $s$ and $\bs{x}$ is lost. Consistency is necessary since centerline curvature depends on $s$ and directly impacts kinematics. 


\begin{IEEEbiography}[{\includegraphics[width=1in,height=1.25in,clip,keepaspectratio]{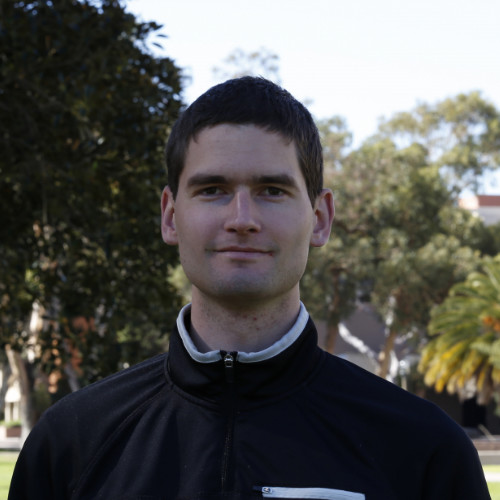}}]%
{Thomas Fork} received the B.Sc. degree in mechanical engineering with highest honors from the University of California at Santa Barbara in 2019. He has been working towards the Ph.D. in mechanical engineering with the University of California at Berkeley under the supervision of Francesco Borrelli since 2020.

His research interests include trajectory optimization, model-predictive control and differential geometry.
\end{IEEEbiography}

\begin{IEEEbiography}[{\includegraphics[width=1in,height=1.25in,clip,keepaspectratio]{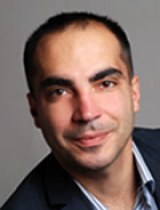}}]%
{Francesco Borrelli}
received the `Laurea’ degree in computer science engineering in 1998 from the University of Naples `Federico II’, Italy. In 2002 he received the PhD from the Automatic Control Laboratory at ETH-Zurich, Switzerland. He is currently a Professor at the Department of Mechanical Engineering of the University of California at Berkeley, USA. He is the author of more than one hundred fifty publications in the field of predictive control. He is author of the book Predictive Control published by Cambridge University Press, the winner of the 2009 NSF CAREER Award and the winner of the 2012 IEEE Control System Technology Award. In 2016 he was elected IEEE fellow. In 2017 he was awarded the Industrial Achievement Award by the International Federation of Automatic Control (IFAC) Council.


His research interests are in the area of model predictive control and its application to automated driving, robotics, food and energy systems.
\end{IEEEbiography}

\end{document}